\documentclass[runningheads]{llncs}
\usepackage{graphicx}
\usepackage{amsmath,amssymb} 
\usepackage{diagbox}
\usepackage{makecell}
\usepackage{hyperref}
\usepackage[percent]{overpic}
\usepackage{multirow}
\usepackage{color} 
\usepackage{wrapfig}
\usepackage{cite}
\usepackage{url}
\usepackage{csquotes}
\usepackage{blindtext}
\newcommand{\orcid}[1]{}

\DeclareMathOperator*{\median}{median}

\def\etal{\emph{et al.} }

\definecolor{orange}{RGB}{255,159,7}
\begin{document}
\pagestyle{headings}
\mainmatter

\def\ACCV20SubNumber{8}  

\title{Self-Guided Multiple Instance Learning for Weakly Supervised Disease Classification and Localization in Chest Radiographs} 
\titlerunning{Self-Guided Multiple Instance Learning}
\authorrunning{C. Seibold et al.}
\index{Seibold,Constantin}
\index{Kleesiek,Jens}
\index{Heinz-Peter,Schlemmer}
\index{Stiefelhagen,Rainer}
\author{Constantin Seibold$^{1,3}$\orcid{0000-0001-6042-8437}, Jens Kleesiek$^2$\orcid{0000-0001-8686-0682},\\ Heinz-Peter Schlemmer$^2$\orcid{0000-0002-9291-0954} and Rainer Stiefelhagen$^1$\orcid{0000-0001-8046-4945}}
\institute{${}{^1}$ {Karlsruhe Institute of Technology}\\ \{constantin.seibold, rainer.stiefelhagen\}@kit.edu\\
${}{^2}$German Cancer Research Center Heidelberg\\
\{j.kleesiek,h.schlemmer\}@dkfz-heidelberg.de\\
${}{^3}$HIDSS4Health - Helmholtz Information and Data Science School for Health, Karlsruhe/Heidelberg, Germany}

\maketitle

\begin{abstract}
\looseness=-1
The lack of fine-grained annotations hinders the deployment of automated diagnosis systems, which require human-interpretable justification for their decision process. In this paper, we address the problem of weakly supervised identification and localization of abnormalities in chest radiographs. To that end, we introduce a novel loss function for training convolutional neural networks increasing the \emph{localization confidence} and assisting the overall \emph{disease identification}. The loss leverages both image- and patch-level predictions to generate auxiliary supervision. 
Rather than forming strictly binary from the predictions as done in previous loss formulations, we create targets in a more customized manner, which allows the loss to account for possible misclassification. We show that the supervision provided within the proposed learning scheme leads to better performance and more precise predictions on prevalent datasets for multiple-instance learning as well as on the NIH~ChestX-Ray14 benchmark for disease recognition than previously used losses.
 \end{abstract}
\section{Introduction}
\looseness=-1
With millions of annually captured images, chest radiographs (\emph{CXR}) are one of the most common tools assisting radiologists in the diagnosing process~\cite{nhs}. The emergence of sizeable CXR datasets such as Open-I or  ChestX-ray14~\cite{openi,bustos2019padchest,johnson2019mimic,irvin2019chexpert,wang2017chestx}, allowed deep Convolutional Neural Networks (\emph{CNN}) to aid the analysis for the detection of pulmonary anomalies~\cite{wang2017chestx,cai2018iterative,shen2018dynamic,tang2018attention,baltruschat2019comparison,rajpurkar2017chexnet,wang2018chestnet,park2019curriculum,rajpurkar2018deep,wang2018low,li2019knowledge,li2019vispi,li2020netnet,wang2018tienet,yan2018weakly,li2018thoracic,liu2019align,rozenberglocalization,guan2018multi}. Despite the success of deep learning,
inferring the correct abnormality location from the network's decision remains challenging. While for supervised tasks, this is achieved through algorithms such as Faster R-CNN~\cite{ren2015faster,liu2016ssd,redmon2018yolov3}, the necessary amount of fine-grained annotation for CXR images to train these models is vastly missing and expensive to obtain. Instead, models are trained using image-level labels parsed from medical reports, which might be inaccurate~\cite{irvin2019chexpert}. As such, the problem of pulmonary pathology identification and localization  is at best weakly supervised.

    \setlength{\abovecaptionskip}{6pt}
    \setlength{\belowcaptionskip}{-25pt}
\begin{figure}[t]
    \centering
    \includegraphics[width=\textwidth,height=0.20\textheight]{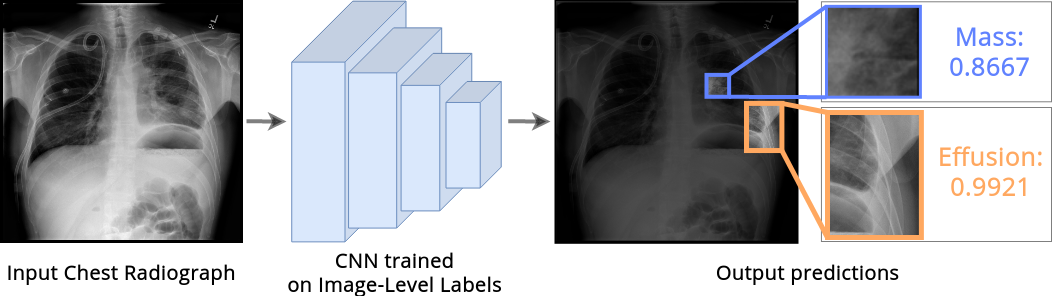}
    \caption{ In our framework, the network reads chest X-ray images and produces overall image-level pathology prediction scores and their corresponding locations.}
    \label{fig:my_label}
\end{figure}
Existing work for weakly-supervised pathology localization in CXR builds either upon network saliency or Multiple-Instance Learning (\emph{MIL}). Saliency-based methods~\cite{wang2017chestx,cai2018iterative,shen2018dynamic,tang2018attention,baltruschat2019comparison,rajpurkar2017chexnet,wang2018chestnet,park2019curriculum} focus primarily on the multi-class recognition task and predict locations implicitly through saliency visualization methods such as CAM, GradCAM, or excitation backpropagation~\cite{zhou2016learning,selvaraju2017grad,zhang2018top}. 
These methods employ global average pooling to merge spatial features during the classification process. However, through this process the CNN makes less indicative decisions, as healthy regions are heavily outweighing the few regions of interest containing the abnormality. 
The other direction combines Fully Convolutional Networks (\emph{FCN}) with MIL to implicitly learn patch-level predictions used for localization~\cite{yan2018weakly,li2018thoracic,liu2019align,rozenberglocalization}. In MIL-based methods, the input data is regarded as a bag of instances where the label is only available on bag-level. The bag will be assigned a positive label if and only if there exists at least one positive instance. This problem formulation fits for diagnosis in medical images as small regions might define the existence of a pathology within the overall image.

In this paper, we focus on MIL-based approaches to diagnose and localize pulmonary abnormalities in CXRs. Much MIL-related work investigated the use of different pooling functions resembling a max-function to aggregate either predictions or embeddings~\cite{wang2017chestx,yao2017learning,ilse2018attention,zhou2017adaptive,mcfee2018adaptive,wang2019comparison,liao2019evaluate,yan2018deep}. 
By balancing all given outputs, networks learn implicitly from the bag label. We argue that this approach overlooks the explicit use of instance-level predictions into training. We present a novel loss formulation split into two stages. While the first stage leads through conventional bag-level classification, the second stage leads to more definitive predictions by generating auxiliary supervision from instance-level predictions. By segregating the prediction maps into foreground, background, and ambiguous regions, the network can provide itself instance-wise targets with differing levels of certainty. 

 The main contributions of our study can be summarized as follows: We provide a novel loss function that applies prediction maps for self-guidance to achieve better classification and localization performance without the necessity to expand a given fully convolutional network architecture. We present the effect of this loss on MIL-specific datasets as well as the ChestX-Ray14 benchmark. The experiments demonstrate competitive results with the current state-of-the-art for weakly supervised pathology classification and localization in CXRs. 

\section{Related Work}
\subsubsection{Automated Chest Radiograph Diagnosis.}
With the release of large-scale CXR datasets~\cite{openi,bustos2019padchest,johnson2019mimic,irvin2019chexpert,wang2017chestx} the development of deep learning-based automated diagnosis methods made noticeable progress in both abnormality identification~\cite{wang2017chestx,cai2018iterative,shen2018dynamic,tang2018attention,baltruschat2019comparison,rajpurkar2017chexnet,wang2018chestnet,park2019curriculum,rajpurkar2018deep,wang2018low,li2019knowledge,li2019vispi,li2020netnet,wang2018tienet,yan2018weakly,li2018thoracic,liu2019align,rozenberglocalization,guan2018multi,chen2019deep,yao2018weakly,guendel2019multi} and the subsequent step of report generation~\cite{li2019vispi,wang2018tienet,liu2019clinically}. However, despite CNNs, at times, surpassing the accuracy of radiologists in detecting pulmonary diseases~\cite{rajpurkar2017chexnet,rajpurkar2018deep}, inferring the correct pathology location remains a challenge due to the lack of concretely annotated data. Initial work such as done by Wang~\etal~\cite{wang2017chestx} or Rajpurkar~\etal~\cite{rajpurkar2017chexnet} uses CAM~\cite{zhou2016learning} to obtain pathology locations.
 Due to the effectiveness and ease of use,  saliency-based methods like CAM became a go-to method for showcasing predicted disease regions~\cite{wang2017chestx,rajpurkar2017chexnet,wang2018chestnet,park2019curriculum,rajpurkar2018deep,yan2018deep}. As such, there exists work to improve CAM visualizations through the use of auxiliary modules or iterative training~\cite{zhang2018self,zhang2018adversarial,wei2017object}.


Alternatively, Li~\etal~\cite{li2018thoracic} propose a slightly modified FCN trained in MIL-fashion to address the problem. Here, each image patch is assigned a likelihood of belonging to a specific pathology. These likelihood-scores are aggregated using a noisy-OR pooling for the means of computing the loss. This approach is extended by Yao~\etal~\cite{yao2018weakly} and Liu~\etal~\cite{liu2019align} who while using different architecture or preprocessing methods stick with the same MIL-based training regime. Similarly, Rozenberg~\etal~\cite{rozenberglocalization} expand Li~\etal's approach through the usage of further postprocessing steps such as the integration of CRFs.

All of these methods approach this task through image-level supervision and try to gain improved localization through changes in architecture, iterative training or postprocessing.  In contrast, rather than modifying a given architecture, we leverage network predictions within the same training step to achieve more confident localization.

\subsubsection{Multiple Instance Learning.}
MIL has become a widely adopted category within weakly supervised learning. It was first proposed for drug activity prediction~\cite{dietterich1997solving} and has since found a use for applications such as sound~\cite{mcfee2018adaptive,wang2019comparison,kong2019sound} and video event tagging~\cite{zhou2017adaptive} as well as weakly supervised object detection~\cite{tang2017multiple,cinbis2016weakly,felipe2020distilling}. While max- and mean- pooling have been common choices for deep MIL networks, recent work investigates the use of the pooling function to combine instance embeddings or predictions to deliver a bag-level prediction~\cite{ilse2018attention,zhou2017adaptive,mcfee2018adaptive,wang2019comparison,liao2019evaluate,yao2018weakly}. 
The choice of pooling function will often resemble the max-operator or an approximation of such to stay in line with common MIL-assumptions. 
Static functions such as Noisy-OR, Log-Sum-Exp or Softmax~\cite{wang2017chestx,li2018thoracic,mcfee2018adaptive} along with learnable ones like adaptive generalized mean, auto-pooling or attention have been proposed~\cite{ilse2018attention,zhou2017adaptive,mcfee2018adaptive}. While the choice of pooling function is a vital part of the overall inference and loss computation in training step in MIL, it in itself does not provide sufficient information as the optimization will still occur only based on the bag-level prediction. In order to accurately impact the training, instance-level predictions are necessary to influence the loss. There exist few methods that leverage the use of artificial supervision within a MIL setting to train the network additionally through instance-level losses~\cite{zhou2017adaptive,morfi2018data,wang2015relaxed,shamsolmoali2020amil}. 
 One direction is to introduce artificial instance-labels for prediction scores above a specified threshold~\cite{wang2015relaxed,zhou2017adaptive,shamsolmoali2020amil}. The loss function splits into a bag-level loss acting in standard MIL fashion by aggregating the predictions and an instance-level prediction where the network gains pixel-wise supervision based on a set prediction threshold~\cite{zhou2017adaptive}. While this approach provides supervision for each instance, it is heavily depending on the initialization potentially introducing a negative bias.
On the other side, Morfi~\etal~\cite{morfi2018data} introduce the MMM loss for audio event detection. This loss provides direct supervision for the extreme values of the bag, whereas the overall bag accumulated using a mean pooling for a bag-level prediction. Despite all instances influencing the optimization, the supervision of this method is limited as it disregards the association for probable positive/negative instances. 

 In an ideal scenario, each positive instance should have a near maximal prediction whereas negative ones should be minimal. However, often, the case presents where the amount of positive bags will sway a classifier towards a biased prediction due to class imbalance. This might lead to all instances within a bag for a certain class to get either high or low prediction scores making strict thresholding difficult to apply. Furthermore, as long as the prediction value distribution within a bag is not separable but rather clumped or uniform existing methods cannot account for a fitting expansion of the decision boundary. In contrast, we adopt instance-level supervision in an adapting way, where the distinctness of the prediction directly defines the influence of the loss.

\looseness=-1
\section{Methodology}
We start this section by defining multiple-instance learning.  We, then, introduce our proposed Self-Guiding Loss~(\emph{SGL}) and how it differs from existing losses. Lastly, we address the use of SGL for classification and weakly supervised localization of CXR pathologies in a MIL setting.
\subsection{Preliminaries of Multiple-Instance Learning}
Assume, we are given a set of bag-of-instances of size $N$ with the associated labels $\mathcal{B}=\{(B_1,y_1), \dots, (B_N,y_N)\}$. 
Let $B_i,i=\{1,\dots,N\}$ be the $i$-th bag-of-instances and $B_{i,j} \in B_i, j\in\{1,\dots,N_i\}$ be the $j$-th instance with $N_i$ being the number of instances of the $i$-th bag.   The associated labels $y_{i} \in \{0,1\}^C$ describe the presence or absence of classes, which can occur independently of each other.  Let $c \in \{1,\dots,C\}$ describe a certain class out of $C$ classes in total.  The label of a bag and an instance  for a specific class $c$ is thus shown by $y^c_{i}\in\{0,1\}$ and $y^c_{i,j}\in\{0,1\}$, respectively. We refer to a target of $1$ as positive and $0$ as negative. The MIL-assumption requires that $y^c_{i}=1$ if and only if there exists at least a single positive instance, hence we can define
\begin{equation}
    y^c_{i} = \max_j y^c_{i,j}.
\end{equation}\label{eq:milAsump}
Note that while the bag-level annotation $y^c_{i}$ is available within the training data,  the instance-level annotation $y^c_{i,j}$ is unknown. 

We aim at learning a classifier to predict the likelihood of each instance in regard to each class within a bag-of-instances. In several works for deep MIL, this classifier might consist of a convolutional backbone $\Psi$ linked with a pooling layer $\Phi$ to combine predictions or features. The class-wise likelihood of a single instance is denoted by $p^c_{i,j}(B_{i,j})\in[0,1]$ with
\begin{equation}
    \begin{gathered}
\mathbf{p}^c_{i}(B_{i})=\{p^c_{i,1}(B_{i,1}),p^c_{i,2}(B_{i,2}),\dots,p^c_{i,N_i}(B_{i,N_i})\}   =  \Psi_c(B_{i})
    \end{gathered}
\end{equation} being the set of all instance-level predictions for class $c$ of the $i$-th bag. 
These instance-level predictions are aggregated using a pooling layer to obtain bag-level predictions 
\begin{equation}
    \begin{gathered}
    p^c_{i}(B_{i}) = \Phi_c(\mathbf{p}^c_{i}(B_{i}))\\
    \end{gathered}
\end{equation}
with $p^c_{i}(B_{i})\in[0,1]$. For brevity, we omit the arguments of the presented functions from this point on.
\begingroup
    \setlength{\abovecaptionskip}{6pt}
    \setlength{\belowcaptionskip}{-15pt}
\begin{figure*}[t]
    \centering
    \includegraphics[width=\textwidth,height=0.2\textheight]{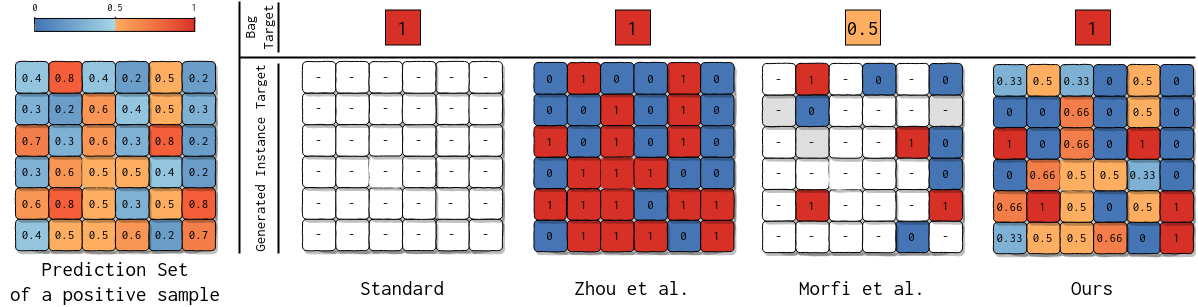}
    \caption{Illustration of supervision for different loss function concepts for MIL. Strict bag-level supervision~(left) provided,  Zhou~\etal's BIL~\cite{zhou2017adaptive}~(center left), Morfi~\etal's MMM~\cite{morfi2018data}~(center right) and on the right our proposed SGL.}
    \label{fig:loss}
\end{figure*}
\endgroup
\subsection{Self-Guiding Loss}
The SGL is designed to address the MIL-setting. Here, one faces an inherent lack of knowledge of the correct instance labels joined with an imbalance between positive and negative instances. Commonly used MIL approaches merge instance predictions and train entirely by optimizing any loss function using the bag's label $y$ and the bag prediction $\mathbf{p}$. This level of supervision is illustrated on the left in Fig~\ref{fig:loss}. The bag label is presented in the top row, while the types of instance supervision are displayed in the bottom. Numbers designate the target label whereas~\enquote{-} denotes no existing supervision for that particular instance. 
While this level of supervision will lead the network to accurate bag-level predictions, inferring the determining instance is not ensured.

Rather than just utilizing the bag, our loss formulation is split in two parts. The first part defines the bag-level loss, while the second part describes how the network's predictions induce artificial supervision to train the network.

\subsubsection{Bag-Level Loss.}
The bag-level loss behaves as in classic MIL approaches. A bag-level prediction is generated by aggregating the network's instance-level predictions. We calculate the loss of this stage using common loss functions $\mathcal{L}$ such as the binary cross-entropy by passing the prediction and target for all classes and bags as follows: 
\begin{equation}\begin{split}\mathcal{L}_{Bag}(\mathcal{B},y) = \frac{1}{C\cdot N} \sum_{c} \sum_{i} \mathcal{L}(p^c_i,y_i)\end{split}\end{equation}\label{eq:1}\noindent
with $i\in\{1,\dots,N\}$ and $c\in\{1,\dots,C\}$. This loss is, hereby, depending on the choice of the pooling function~$\Phi$ and provides leeway for the instance-level loss to step in. 
\subsubsection{Instance-Level Loss.} To outline the instance-level loss, we start with the assumption that a network trained just from bag labels will inevitably assign some positive instances a noticeably higher prediction score than most negative instances. From this, we derive three types of instance predictions. Instances with a high score are likely to be considered positive, whereas instances with a low score as negative. Instances with scores close to the decision boundary are rather ambiguous as they may easily be swayed in the course of training and as such do not pose an as concrete implication about the actual class of the instance. Pursuing this line of thought we establish three types of supervision  based on the certainty level of each prediction within a bag. 

 Our first step is to normalize the prediction set using the common min-max feature scaling. We apply this to avoid cases of biases stemming from either algorithmic decisions such the choice of the pooling function or general data imbalance. We denote the resulting rescaled bag of predictions $\theta$ by
\begin{equation}
    \theta^c_{ij}=\frac{p^c_{ij}-\min(\mathbf{p}^c_{i})}{\max(\mathbf{p}^c_{i})-\min(\mathbf{p}^c_{i})}
\end{equation}\label{eq:2}\noindent
with min and max being functions returning the minimal and maximal values within a set respectively. 
The normalized predictions are then used within a ternary mask $M$ depicting targets stemming from the previously named cases similar to Hou~\etal~\cite{Hou2018Self} and Zhang~\etal~\cite{zhang2018self}. For this, we define a higher and lower threshold to partition the prediction set, $\delta_h$ and $\delta_l$ respectively with $\delta_h+\delta_l =1$ and $\delta_h\geq\delta_l\geq0$. Everything larger than the upper threshold $\delta_h$ will be regarded as a positive instance and all instances with scores lower than $\delta_l$ as negative. The target mask $M$ is then defined for each instance $j$ in the bag $i$ for class $c$ by
\begin{equation}
M^c_{i,j} = 
\begin{cases}
     0 & \text{, if } \theta^c_{i,j} < \delta_l \text{ or } y_i^c=0   \\
      \theta^c_{i,j} & \text{, if } \delta_l \leq \theta^c_{i,j} \leq \delta_h\\
      1 & \text{, if } \delta_h < \theta^c_{i,j}\\
\end{cases}  .
\end{equation}\label{eq:3} 
For distinctly positive and negative predictions, we obtain instance-wise supervision with a target value of 1 and 0 respectively. We can also presume based on Eq.~1. 
that each instance within negative bags is also negative. Thus, we can set all values of their masks to $0$. The remaining uncertain regions, however, do not allow for as explicit label assignment. While we want to enforce the networks decision process, we also have to account for possible missassignment. Thus, rather than setting a fixed target value, we set the target to be $\theta$. This process shows some similarity to the popular label smoothing procedure~\cite{szegedy2016rethinking}. Rather than using maximal valued targets, the maps adjusted value is inserted into the loss function as target value. This slightly pushes the loss into the direction of the most extreme predictions within the uncertain instance set. By doing so we steadily increase the amount distinctly positive and negative predictions over the course of training.

We can construct the loss using a fundamental loss function $\mathcal{L}$ like binary cross entropy by utilizing $M$ as target. The instance-level loss is then defined as 
\begin{equation} 
    \mathcal{L}_{Inst}(\mathcal{B},M) =  \sum_i\sum_c\sum_j2^{\alpha^c_i-1}\cdot\mathcal{L}(p^c_{i,j},M^c_{i,j}),
\end{equation}\label{eq:4}
where each part is being normalized by the number of pixels with the respective supervision types.
This way, we strengthen the networks decision process for its more certain instances. 
We, further, consider a weighing factor $\alpha$ to influence the bag's impact based on the overall certainty of its prediction. We define $\alpha$ by 
\begin{equation}
    \alpha^c_i = \max(\max(\mathbf{p}_{i}^c) -\median(\mathbf{p}_i^c)),1-y_i)
\end{equation}
Since a positive bag in a common MIL setting should have a low valued median due to a limited amount of positive instances, it is weighted highly if the network is able to clearly separate positive from negative predictions. Thus, for positive bags, $\alpha=0$ if all predictions result in the same value and  $\alpha=1$ if the network is able to clearly separate positive from negative instances under the assumption that the number of positive instances is vastly smaller than the number of negative ones. For negative bags, $\alpha=1$ holds due to the given supervision. 

The complete loss is then defined by \begin{equation}\mathcal{L}_{SGL}(p_i,y_i) =
\mathcal{L}_{Bag}
+\lambda \cdot \mathcal{L}_{Inst}, 
\end{equation} 
with $\lambda$ denoting the weighing hyperparameter of the instance-level loss.

An example of the final supervision for our loss is displayed in Fig~\ref{fig:loss}. The standard approach on the left uses no instance-level supervision. In the center left, Zhou~\etal's BIL provides a positive label for each instance above the $0.5$ threshold and a negative else, while maintaining the bag supervision. The MMM loss by Morfi~\etal, in the center right, considers positive labels for the maximum instances and negative ones for minimal instances. It further uses the target of $0.5$ for a mean pooled prediction. Opposed to this, our loss adapts its assumed supervision to the produced predictions. Rather than just using the maximum or applying set thresholding, we threshold on a rescaled set of predictions, thus avoiding a common problem occurring with imbalanced data. Our formulation incorporates all instance predictions while providing a margin of error based on the networks certainty over the smoothed targets $\theta$ and the weighing factor $\alpha$. 
\begingroup
    \setlength{\abovecaptionskip}{0pt}
    \setlength{\belowcaptionskip}{-16pt}
\begin{figure}[t]
    \centering
    \includegraphics[width=\linewidth, height=0.20\textheight]{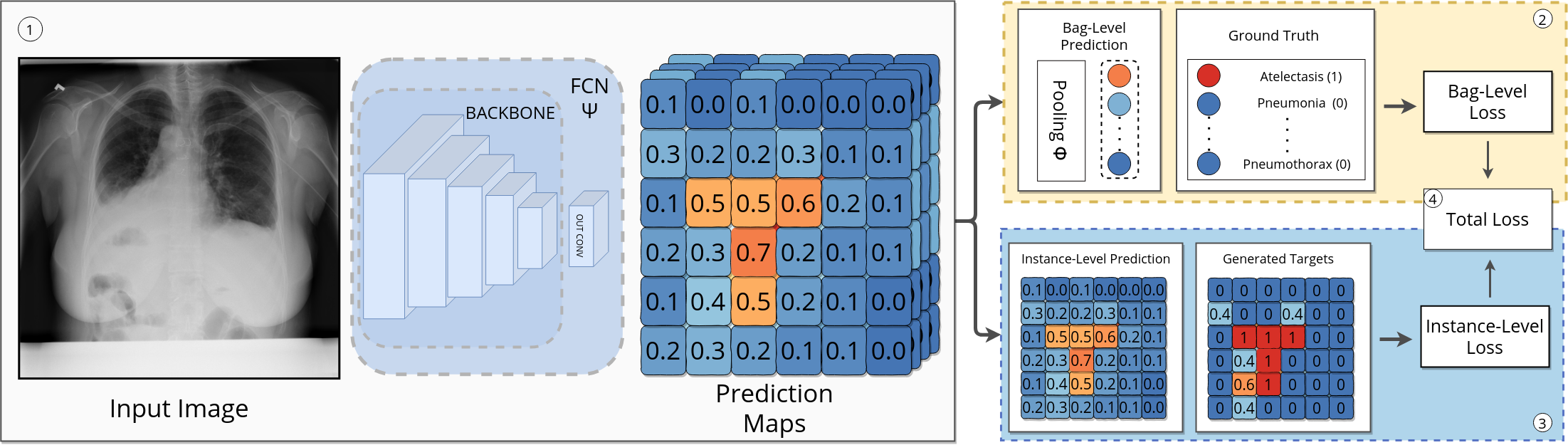}
    \caption{Overview of considered framework for thoracic disease identification and localization. A chest X-ray is passed through an FCN and produces a prediction map. The prediction map is used to compute the instance- and bag-level losses.
    }
    \label{fig:overview}
\end{figure}
\endgroup
\subsection{MIL for Chest Radiograph Diagnosis}
We consider a MIL scenario for CXR diagnosis. We build upon the assumption that singular patches (instances) of an image (bag-of-instances) can infer the occurrence of such a pathology (class).  An example of this is the class \enquote{nodules}, which can take up minimal space within the image. We are given just image-level labels for pathologies, while more detailed information such as bounding box or pixel-level supervision remains hidden. The bag is associated with a class if and only if there exists at least one instance causing such implication. The goal is to learn a model that when given a bag of instances can predict the bag's label based on the instance information. By classifying the bag's instances the model provides insight regarding which regions are affected by a pathology.

\noindent\textbf{Overview.}
In Figure~\ref{fig:overview}, we illustrate an overview of the considered scheme for CXR diagnosis. Firstly, an FCN processes CXR-images, which results in patch-wise classification scores for each abnormality.  The number of patches stems from their perceptual field, which is a result of backbone architecture. Each patch is independently processed via a $1\times1$ convolutional classification layer. In this work, we do not add specific modules to our backbone network.  These patch predictions are aggregated in the second part using a pooling layer, which produces a bag-level prediction for our bag-level loss.  The third part applies an instance-level loss function based on the patch-wise predictions.  In the fourth part, both the instance and bag-level losses join for optimization. Here, we further penalize the occurrence of non-zero elements in $M$ using an $L_2$-Norm.

\noindent\textbf{Choice of Pooling Function.}
The choice of the correct pooling function is vital for any MIL-setting to produce accurate bag-level predictions. 
Methods like max and mean pooling will lead to imprecise decisions. In the context of MIL in CXR diagnosis, Noisy-OR found use, but this function suffers from the numerical instability stemming from the product of a multitude of instances. Rather than letting singular instances influence the decision process, we choose to employ the Softmax-pooling, which has found success in audio event detection~\cite{mcfee2018adaptive,wang2019comparison}. It provides a meaningful balance between instance-level predictions to let each instance influence the bag level loss based on its intensity.

\looseness=-1
\section{Experiments}
\subsection{Datasets}
\noindent\textbf{MNIST-Bags.}
In a similar fashion to Ilse~\etal~\cite{ilse2018attention}, we use the MNIST-bags~\cite{ilse2018attention,lecun1998gradient} dataset to evaluate our method for a MIL-setting. A bag is created grayscale MNIST-images of size $28\times28$, which are resized to $32\times32$. A bag is considered positive if it contains the label \enquote{9}. The number of images in a bag is Gaussian-distributed based on a fixed bag size. We investigate different average bag sizes and amounts of training bags.  During evaluation 1000 bags created from the MNIST test set of the same bag size as used in training. We average the results of ten training procedures.\footnote{\url{www.github.com/ConstantinSeibold/SGL}}


\noindent\textbf{CIFAR10-Bags.}
We build CIFAR10-bags from CIFAR10~\cite{krizhevsky2009learning} in a similar fashion to MNIST-bags.  We choose to create 2500 and 5000 training and test bags respectively with fixed bag sizes.  A bag here is considered positive if it contains the label \enquote{dog}. We investigate in these experiments the influence of a varying number of positive instances per bag. We average  five training runs.

\noindent\textbf{NIH ChestX-ray14.}
To present the effect of our loss for medical diagnosis, we conduct experiments on the NIH ChestX-ray14 dataset~\cite{wang2017chestx}. It contains 112,120 frontal-view chest X-rays taken from 30,805 patients with 14 disease labels. Unless further specified, we resize the original image size of $1024\times1024$ to $512\times512$. We use the official split between train/val and test, as such we get a 70\%/10\%/20\% split. Also, 880 images with a total of 984 images with bounding boxes for 8 of the 14 pathologies from the test set.

\subsection{Implementation Details}
For all MNIST-Bags-experiments, we use a LeNet5 model~\cite{lecun1998gradient} as Ilse~\etal~\cite{ilse2018attention}. We apply max-pooling, $\delta_l=0.3$ and $\lambda=1$ for our method unless further specified. We train BIL~\cite{zhou2017adaptive} using mean-pooling as we found it unable to train with max-pooling.

For all CIFAR10-bags-experiments, we train a ResNet-18~\cite{he2016deep} with the same optimizer hyperparameters and batchsize of 64 for 50 epochs. We apply max-pooling, $\delta_l=0.3$ and $\lambda=1$ for our method.

For the experiments on NIH ChestX-ray14, each network is initialized using an Image-Net pretraining. We use the same base model as Wang~\etal~\cite{wang2017chestx} by employing a ResNet-50~\cite{he2016deep}. We replace the final fully connected and pooling layers with a convolutional layer of kernel size $1\times1$, resulting in the same number of parameters as Wang~\etal We follow standard image normalization~\cite{russakovsky2015imagenet}. For training, we randomly crop the images to size 7/8-th of the input image size, whereas we use the full image size during test time.  We train the network for 20 epochs using the maximum batch-size for our GPU using Adam~\cite{kingma2014adam} with a learning rate, weight decay, $\beta_1$ and $\beta_2$ of $10^{-4},10^{-4},0.9$ and $0.999$ respectively. We decay the learning rate by $0.1$ every 10 epochs. We set $\delta=0.3$ and $\lambda=20$. We increase $\lambda$ to keep the two losses on similar magnitudes. The model is implemented using Pytorch~\cite{paszke2017automatic}.
\subsection{Evaluation Metrics}
We evaluate the classification ability of our network via the area under the ROC-curve
(\emph{AUC}). 
To evaluate the localization ability we apply average intersection-over-union (\emph{IoU}) to calculate the class-wise localization accuracy similar to Russakovsky~\etal~\cite{russakovsky2015imagenet}. To compute the localization scores, we threshold the probability map at the scalar value $T_p$ to get the predicted area and compute the intersection between predicted and ground truth area to compute the IoU. In the case of MNIST- and CIFAR10-bags IoU is computed as the intersection between predicted positive instances and ground truth positive instances at $T_p=0.5$. The localization accuracy is calculated by
$\frac{\#hit}{\#hit+\#miss}$, 
where an image has the correct predicted localization ($hit$) iff it
has the correct class prediction and a higher overlap than a predefined threshold $T_{IoU}$.
    \setlength{\abovecaptionskip}{0pt}
    \setlength{\belowcaptionskip}{-18pt}
\begin{figure}[t]
    \centering
    \begingroup
    \setlength{\tabcolsep}{-5pt} 
    \begin{tabular}{ccc}
         \includegraphics[width=0.37\textwidth, height=0.18\textheight]{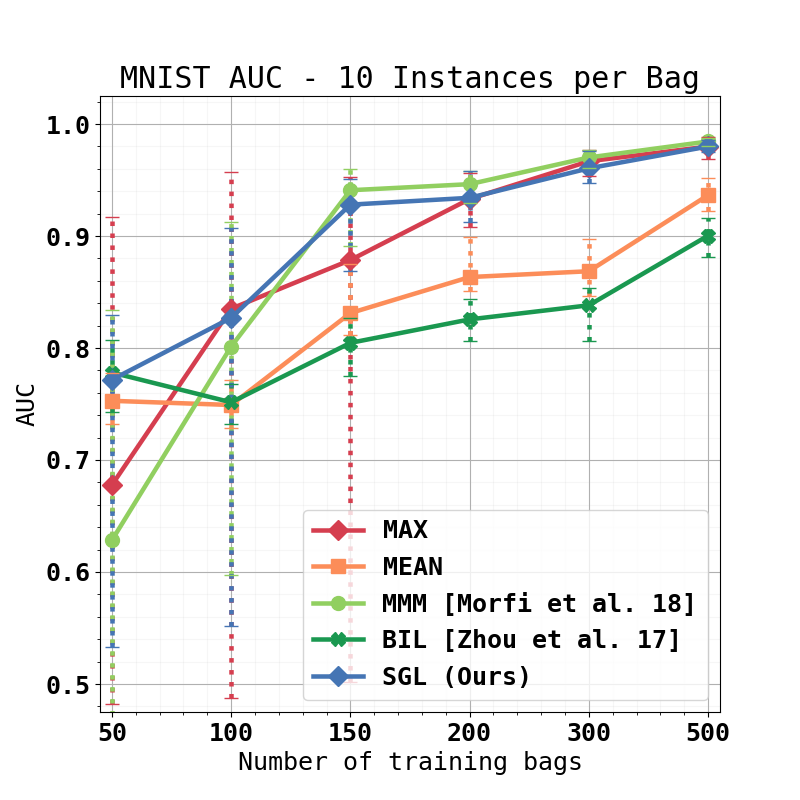}&
         \includegraphics[width=0.37\textwidth, height=0.18\textheight]{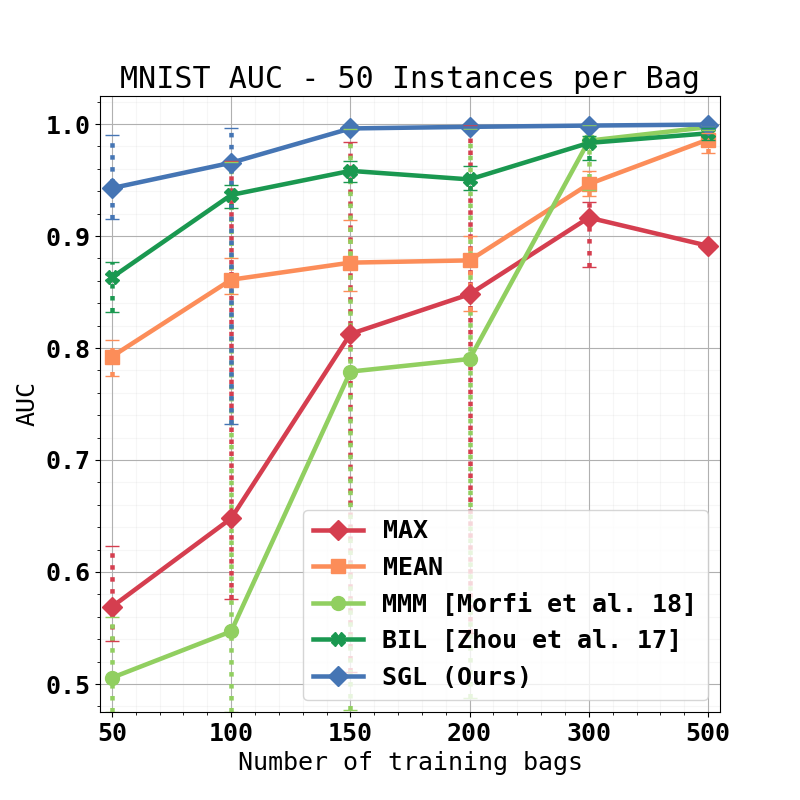}&
         \includegraphics[width=0.37\textwidth, height=0.18\textheight]{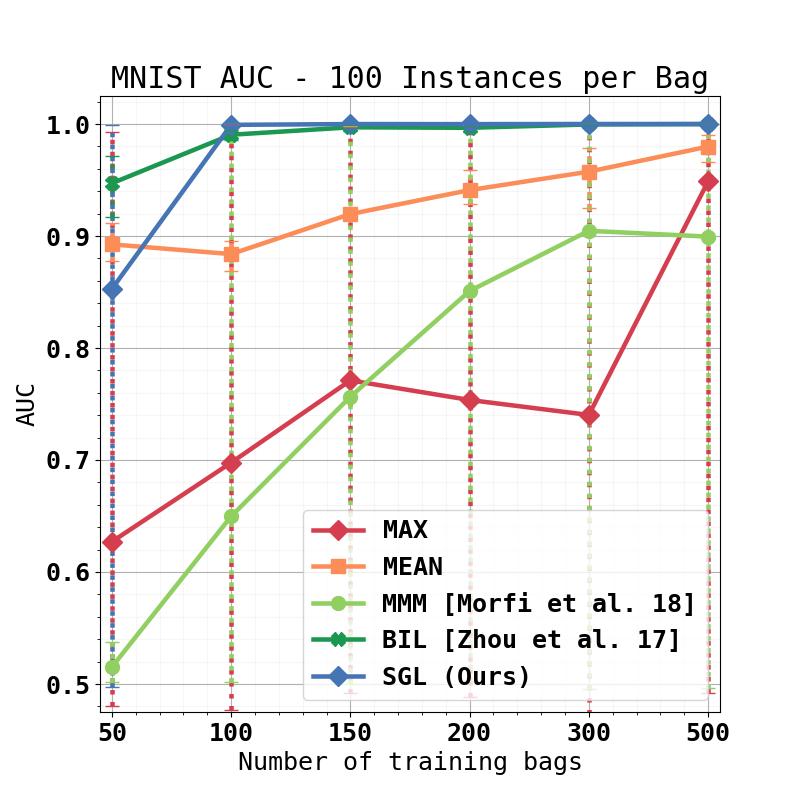} \\
         \includegraphics[width=0.37\textwidth, height=0.18\textheight]{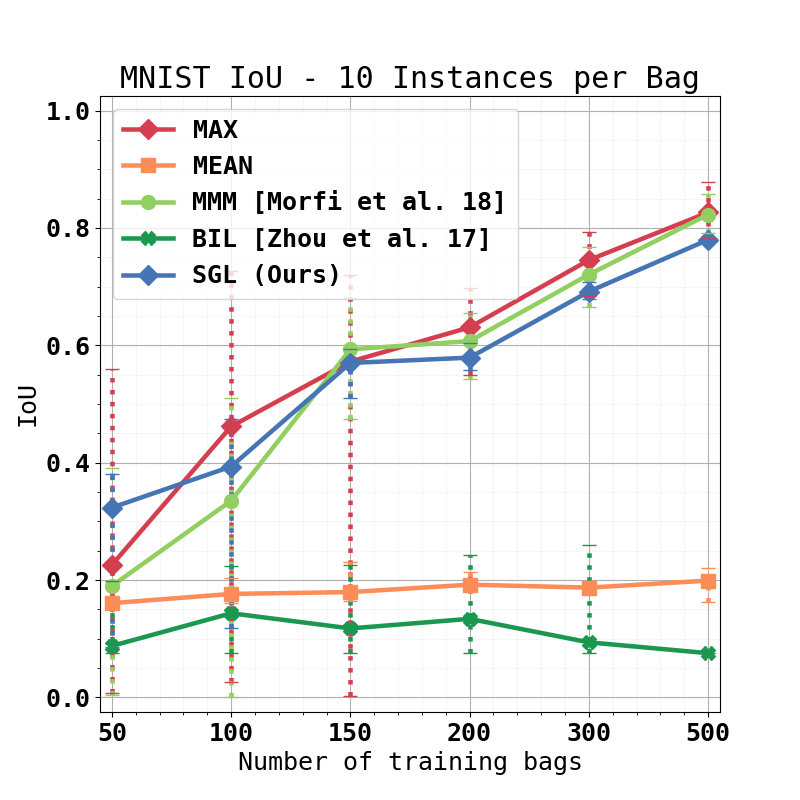}&
         \includegraphics[width=0.37\textwidth, height=0.18\textheight]{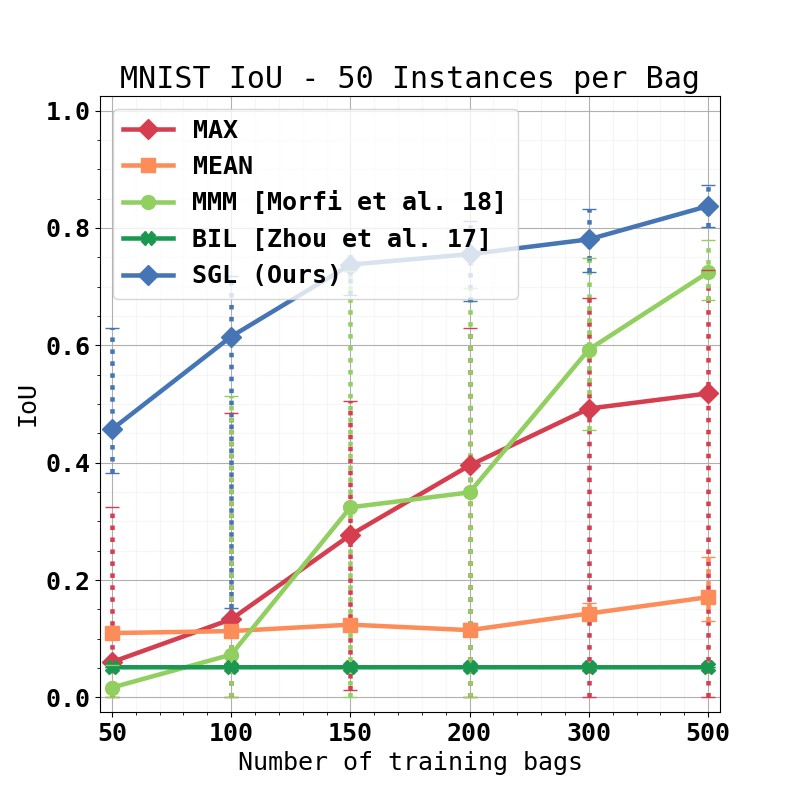}&
         \includegraphics[width=0.37\textwidth, height=0.18\textheight]{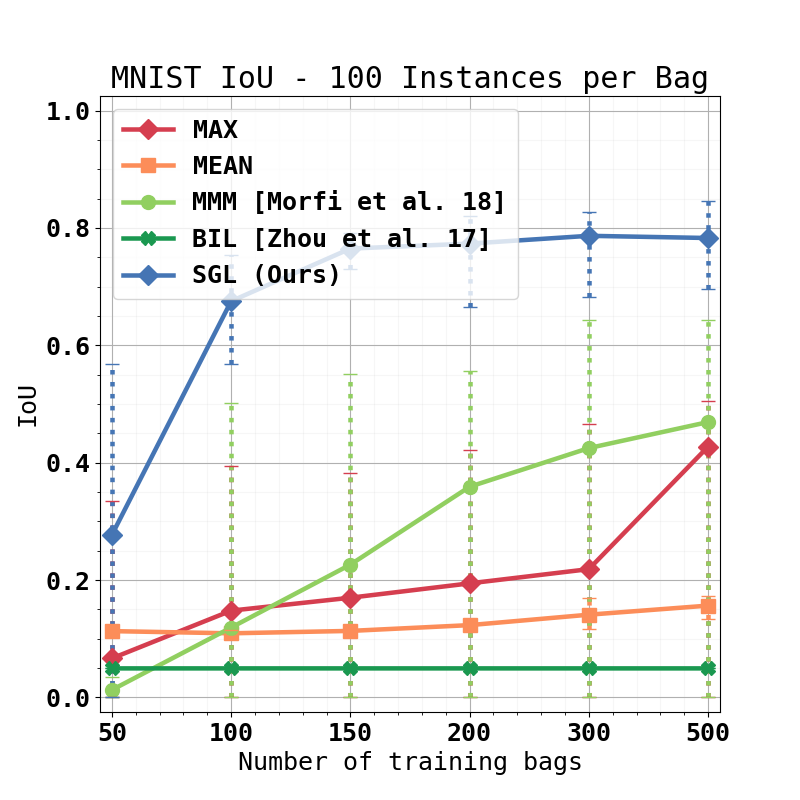}
    \end{tabular}
    \endgroup
    \caption{Test AUC and IoU for MNIST-Bags for differing avg. instances per bag.}
    \label{fig:mnist_quant}
\end{figure}
    \setlength{\abovecaptionskip}{0pt}
    \setlength{\belowcaptionskip}{-16pt}
\begin{figure}[t]
    \begingroup
    \setlength{\tabcolsep}{-0pt} 
    \renewcommand{\arraystretch}{-0.5}
    \centering
    \begin{tabular}{cc}
    \centering
            \begin{tabular}{c}
                 \includegraphics[width=0.5\textwidth,height=0.1625\textheight]{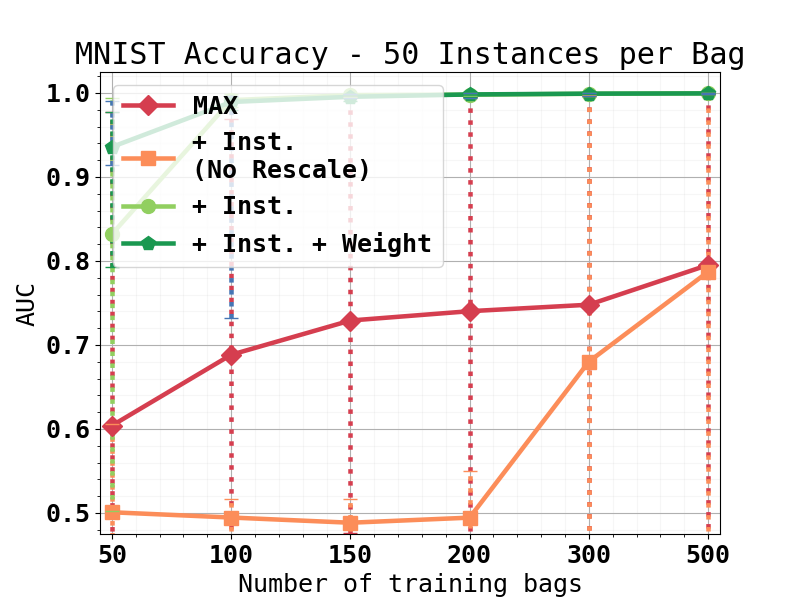}
                    \\
                 \includegraphics[width=0.5\textwidth,height=0.1625\textheight]{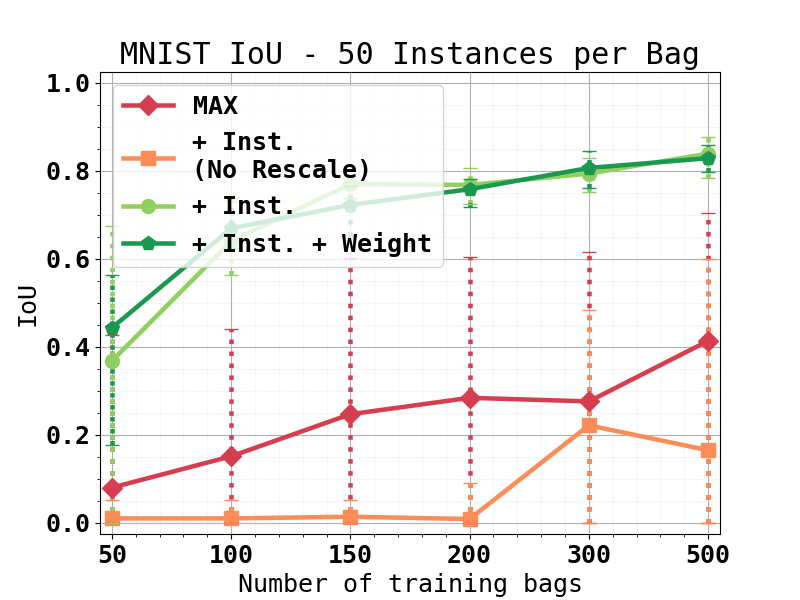}\\(a)
            \end{tabular} 
        &  
            \begin{tabular}{c}
                 \includegraphics[width=0.5\textwidth,height=0.1625\textheight]{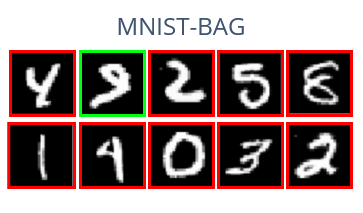}
                    \\
                 \includegraphics[width=0.5\textwidth,height=0.1625\textheight]{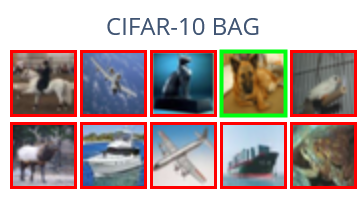}  \\ (b)
                 
    \end{tabular}    \end{tabular}
    \endgroup

    \caption{Ablation study of SGL in Figure (a). Figure (b) displays exemplary positive bags of size 10 for MNIST-bags (top) and CIFAR10-bags (bottom). Positive instances denoted in \textcolor{green}{green}, negative ones in \textcolor{red}{red}.}
    \label{fig:abl}
\end{figure}

\subsection{Results}

\noindent\textbf{MNIST-Bags.}
The AUC and IoU results for the mean bag sizes of 10, 50, and 100 with a varying number of given training bags are displayed from left to right in the top and bottom row of Fig.~\ref{fig:mnist_quant}. We present the average of the runs as well as the best and worst runs for each method. For small bags, our method performs similarly to the simple max-pooling in both AUC and IoU. We attribute this average performance to the small number of instances in a bag, which does not allow to make proper use of our ternary training approach. As we increase the bag size to 50 and 100 our proposed loss performs better than the max-pooling baseline but also than the other methods for both metrics. We can see the difference notably in the IoU, where our loss achieves nearly double the performance of the next best method for almost all amounts of training bags. We notice that our approach does not pose a trade-off between confident predictions and overall AUC but manages to facilitate a training environment which improves both metrics. It is also worth mentioning that while increasing the amount of training bags improves the method for any bag size our loss achieves exceptional performance for both AUC and IoU with a relatively small number training examples for larger bag sizes. We can reason that the further use of self-guidance can potentially improve a method regardless of dataset size.

In Fig.~\ref{fig:abl} (a), we present ablation studies involving different constellations of the loss. When considering the loss components, we start with just max-pooling baseline and successively add parts of SGL. Max-pooling alone struggles with the identification of positive/negative bags, however, improves slowly in terms of IoU and AUC with increasing numbers of training bags. When adding the proposed loss without the rescaling mentioned in Eq. \ref{eq:2} and weighting component (shown by \emph{Inst.(No Rescale)}) the method becomes incapable to learn as even random initializations might skew the network towards incorrect conclusions. When adding the rescaling component (shown by \emph{Inst.}) the model drastically outperforms prior parts in both metrics. Doing so achieves higher maximums than with the applied weighting factor $\alpha$ displayed by \emph{Inst.+Weight}. However, the addition of the weighting factor provides a more stable training, specifically for smaller amounts of training data. 

\noindent\textbf{CIFAR10-Bags.}
    \setlength{\abovecaptionskip}{6pt}
    \setlength{\belowcaptionskip}{-10pt}
\begin{figure}[t]
    \begingroup
    \setlength{\tabcolsep}{-8pt} 
    \renewcommand{\arraystretch}{1.05}
    \centering
    \begin{tabular}{cc}
         \includegraphics[width=0.56\textwidth,height=0.19\textheight]{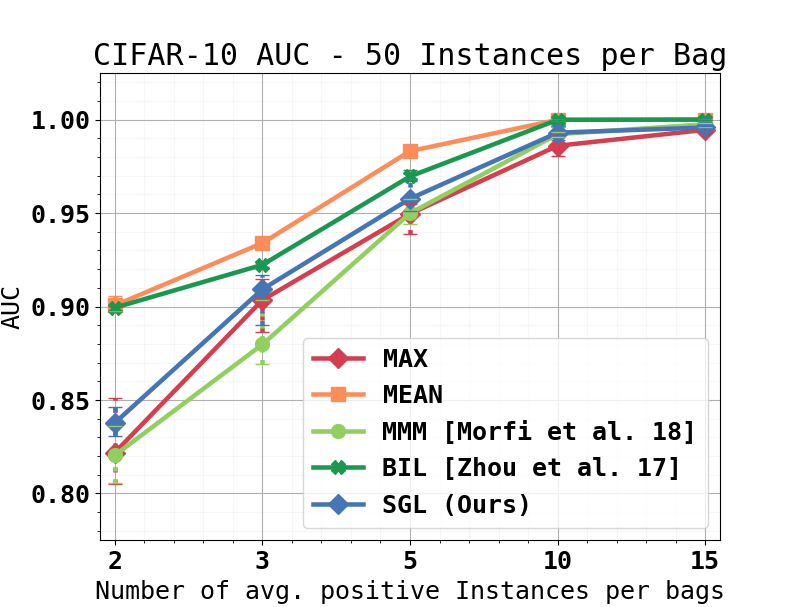}&\includegraphics[width=0.56\textwidth,height=0.19\textheight]{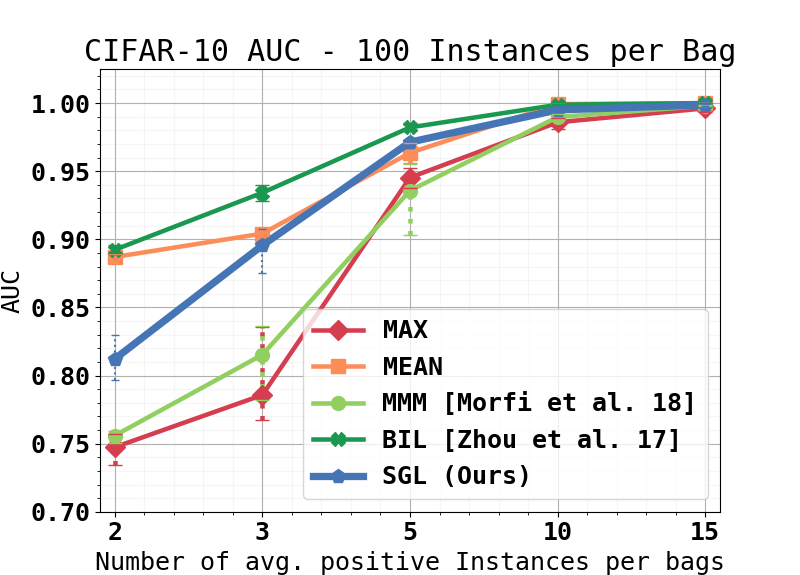}\\
         \includegraphics[width=0.56\textwidth,height=0.19\textheight]{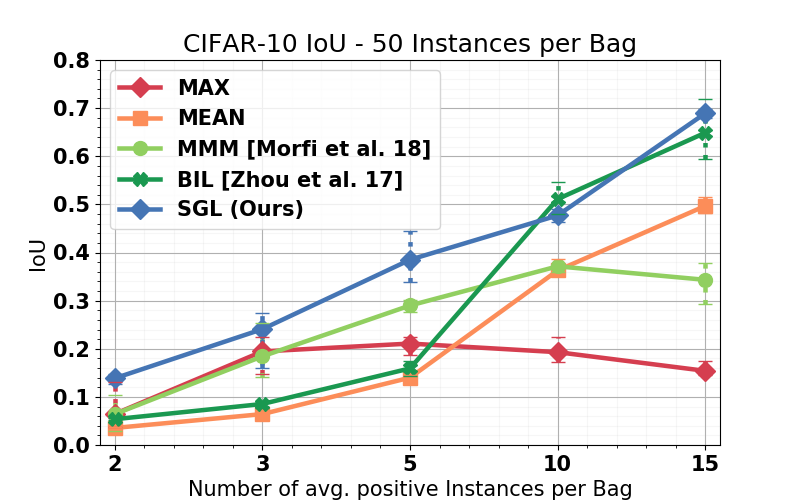}&\includegraphics[width=0.56\textwidth,height=0.19\textheight]{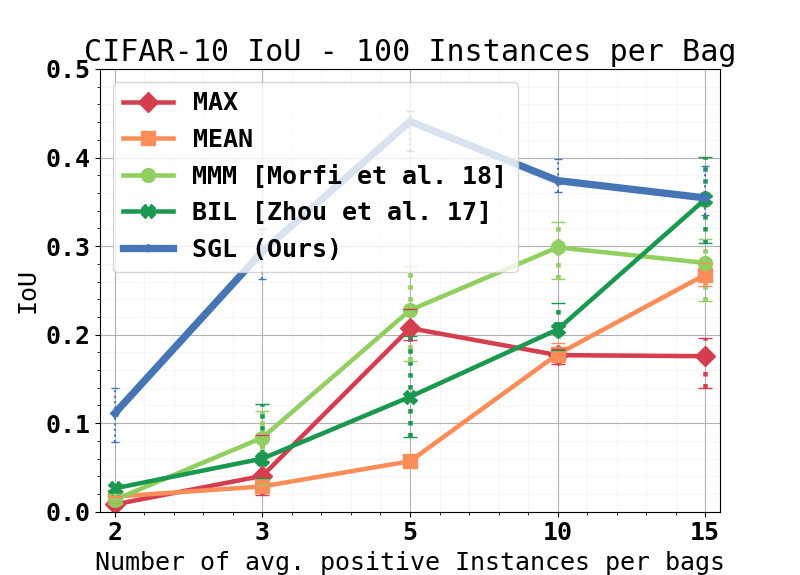}
    \end{tabular}
    \endgroup
    \caption{Test AUC and IoU for CIFAR10-Bags for differing number of average positive instances per bag with a bag size of 50 and 100.}
    \label{fig:cifar}
\end{figure}
The AUC and IoU results for the mean bag sizes of 50 and 100 with a varying number of positive instances per bag are in the top and bottom row of Fig.~\ref{fig:cifar}. For smaller bag sizes, we observe that straight forward mean-pooling achieves the best AUC scores for CIFAR10-Bags. Overall SGL improves over straight forward max-pooling for any number of instances. In regards to IoU, our method manages to outperform other methods for nearly any number of positive instances per bag. For larger bag sizes SGL achieves roughly the same performance as BIL which trained using mean-pooling in terms of AUC while outperforming it in IoU for all amounts of positive instances per bag. The addition of self-guidance manages to bridge shortcomings of max-pooling, boosting its classification accuracy for any bag size or number of positive instances.
\setlength{\abovecaptionskip}{10pt}
\setlength{\belowcaptionskip}{-15pt}
\begin{table}[b]
    \centering
    \resizebox{\textwidth}{!}{%
    \begin{tabular}{|c||c|c|c|c|c|c|c|c|c|c|c|c|c|c|c|}
    \hline
        \diagbox[width=7.5em]{Method}{Pathologies}& At.  & Card.  &  Cons. &  Ed. & Eff.  & Emph.  & Fib. & Hernia  & Inf.  &  Mass &  Nod. & \makecell{Pl. \\ Th.}   & Pn.  & Pt.  & Mean     \\\hline\hline
         Wang~\etal&0.70&0.81&0.70&0.81&0.76&0.83&0.79&0.87&0.66&0.69&0.67&0.68&0.66&0.80& 0.75 \\\hline
         Li~\etal*&\textbf{0.80}&0.87&\textbf{0.80}&0.88&0.87&0.91&0.78&0.70&0.70&0.83&0.75&0.79&0.67&0.87& 0.81\\\hline
         Liu~\etal*&0.79&0.87&0.79&\textbf{0.91}&\textbf{0.88}&0.93&0.80&0.92&0.69&0.81&0.73&0.80&\textbf{0.75}&0.89& \textbf{0.83} \\\hline\hline
         ResNet-50+SGL&0.78&\textbf{0.88}&0.75&0.86&0.84&\textbf{0.95}&\textbf{0.85}&\textbf{0.94}&\textbf{0.71}&\textbf{0.84}&\textbf{0.81}&\textbf{0.81}&0.74&\textbf{0.90}& \textbf{0.83} \\\hline
    \end{tabular}%
    }
    \caption{Comparison of classification performance for CXR pathologies on the NIH ChestX-Ray14 dataset. Here, 70\% of all images were used for training with no bounding box annotations available. Evaluations were performed on the official test split containing 20\% of all images. \enquote{*} denotes usage of additional bounding box supervision.}
    \label{tab:class}
\end{table}
\setlength{\abovecaptionskip}{10pt}
\begin{table}[t]
    \centering
    \resizebox{\textwidth}{!}{%
    \begin{tabular}{|c|c|c|c|c|c|c|c|c|c|c|}
    \hline
        $T_{IoU}$ & Model & Atelectasis & Cardiomegaly & Effusion & Infiltration & Mass & Nodule & Pneumonia & Pneumothorax & Mean\\\hline\hline
        
         \multirow{4}{*}{0.1}&Wang~\etal\cite{wang2017chestx}&0.69&0.94&0.66&0.71&0.40&0.14&0.63&0.38&0.57 \\\cline{2-11}
         &Li~\etal ~\cite{li2018thoracic}*&\textbf{0.71}&\textbf{0.98}&\textbf{0.87}&\textbf{0.92}&\textbf{0.71}&0.40&0.60&\textbf{0.63}&\textbf{0.73} \\\cline{2-11}
         &Liu~\etal~\cite{liu2019align}&0.39&0.90&0.65&0.85&0.69&0.38&0.30&0.39&0.60 \\\cline{2-11}
         &SGL (Ours)&0.67&0.94&0.67&0.81&\textbf{0.71}&\textbf{0.41}&\textbf{0.66}&0.43&0.66 \\\hline\hline
         \multirow{4}{*}{0.3}&Wang~\etal\cite{wang2017chestx}&0.24&0.46&0.30&0.28&0.15&0.04&0.17&0.13&0.22 \\\cline{2-11}&Li~\etal ~\cite{li2018thoracic}*
         &\textbf{0.36}&\textbf{0.94}&\textbf{0.56}&\textbf{0.66}&0.45&\textbf{0.17}&\textbf{0.39}&\textbf{0.44}&\textbf{0.50} \\\cline{2-11}&Liu~\etal~\cite{liu2019align}
         &0.34&0.71&0.39&0.65&\textbf{0.48}&0.09&0.16&0.20&0.38 \\\cline{2-11}
         &SGL (Ours)&0.31&0.76&0.30&0.43&0.34&0.13&\textbf{0.39}&0.18&0.36 \\\hline\hline
         \multirow{4}{*}{0.5}&Wang~\etal\cite{wang2017chestx}&0.05&0.18&0.11&0.07&0.01&0.01&0.01&0.03&0.06 \\\cline{2-11}&Li~\etal ~\cite{li2018thoracic}*
         &0.14&\textbf{0.84}&\textbf{0.22}&0.30&0.22&0.07&\textbf{0.17}&\textbf{0.19}&\textbf{0.27} \\\cline{2-11}&Liu~\etal~\cite{liu2019align}
          &\textbf{0.19}&0.53&0.19&\textbf{0.47}&\textbf{0.33}&0.03&0.08&0.11&0.24 \\\cline{2-11}
         &SGL (Ours)&0.07&0.32&0.08&0.19&0.18&\textbf{0.10}&0.12&0.04&0.13\\ \hline\hline
         \multirow{4}{*}{0.7}&Wang~\etal\cite{wang2017chestx}&0.01&0.03&0.02&0.00&0.00&0.00&0.01&0.02&0.01 \\\cline{2-11}&Li~\etal ~\cite{li2018thoracic}*
         &0.04&\textbf{0.52}&0.07&0.09&0.11&\textbf{0.01}&\textbf{0.05}&0.05&0.12 \\\cline{2-11}&Liu~\etal~\cite{liu2019align}
         &\textbf{0.08}&0.30&\textbf{0.09}&\textbf{0.25}&\textbf{0.19}&\textbf{0.01}&0.04&\textbf{0.07}&\textbf{0.13} \\\cline{2-11}
        &SGL (Ours)&0.02&0.01&0.1&0.00&0.04&0.00&0.03&0.01&0.01 \\\hline

    \end{tabular}
    
    }
    \caption{Disease localization accuracy are evaluated with a classification threshold of 0.5. \enquote{*} denotes additional bounding box supervision.}
    \label{tab:abl_localization_results}
\end{table}


\noindent\textbf{NIH ChestX-Ray 14: Multi-Label Pathology Classification.}
Table~\ref{tab:class} shows the AUC scores for all the disease classes.  We compare the results of our loss function with a common classification approach by Wang~\etal~\cite{wang2017chestx}, the MIL-based methods proposed by Li~\etal\cite{li2018thoracic} and Liu~\etal~\cite{liu2019align}. The latter two employ noteworthy architectural adaptations and train using bounding box supervision. All of the named methods utilize a ResNet-50 as backbone network. We outperform the baseline ResNet-50 of Wang~\etal in all categories. We observe that our loss formulation achieves better classification performance than all other methods in \textbf{9} of \textbf{14} classes in total. We also reach a better mean performance than other methods, which use further bounding box annotations and architectural modifications such employing additional networks~\cite{liu2019align} or further convolutional layer~\cite{li2018thoracic,liu2019align}.

\noindent\textbf{NIH ChestX-Ray 14: Pathology Localization.}
We evaluate the localization ability of the prior named methods through the accuracy over an IoU threshold. For our method, we upsample each prediction map using Nearest-Neighbor-Interpolation. We construct bounding boxes around the connected component of the maximum prediction after applying common morphological operations. The results are displayed in Table~\ref{tab:abl_localization_results} for the IoU thresholds $T_{IoU} \in~\{0.1,0.3,0.5,0.7\}$. We, further, display qualitative examples for each pathology in Figure~\ref{fig:pathologies}. For the visualization, we use no morphological postprocessing. We compare our method against a baseline version of our model trained only using the bag-level loss with a mean-pooling function. The expert annotation is displayed by a green bounding box, while the predicted one is orange.

Our method achieves favourable performance across all pathologies on a threshold of $T_{IoU}=0.1$. It generally outperforms the baseline of Wang~\etal~\cite{wang2017chestx}. For higher thresholds, our model falls behind the more specified approaches of Li~\etal~\cite{li2018thoracic} and Liu~\etal~\cite{liu2019align}. We ascribe the suboptimal quantitative performance to the factors of low spatial output resolution, which can hinder passing the IoU threshold especially for naturally small classes such as \emph{Nodules},  and the overall coarse annotation as can be seen in Figure~\ref{fig:pathologies} e.g. the pathology \emph{Infiltrate}. Here, an infiltrate affects the lung area, which the model correctly marks, yet the bounding box naturally includes the cardiac area, thus diminishing the IoU. 

In Figure~\ref{fig:pathologies}, we see that our proposed method can generally make more precise predictions compared to the baseline model. Furthermore, the model can more distinctly separate between healthy and abnormal tissue. These results indicate the ability of our loss to lead itself towards more refined predictions.
\begingroup
\renewcommand{\arraystretch}{0.35} 
\setlength{\abovecaptionskip}{2pt}
\setlength{\belowcaptionskip}{-15pt}
\begin{figure}[t]

    \centering
    \begin{tabular}{cc}
        \begin{tabular}{ccc}
                \multicolumn{3}{c}{\fontfamily{lmtt}\selectfont\scriptsize Pneumothorax}\\
              \frame{\includegraphics[width=0.155\textwidth,height=0.155\textwidth]{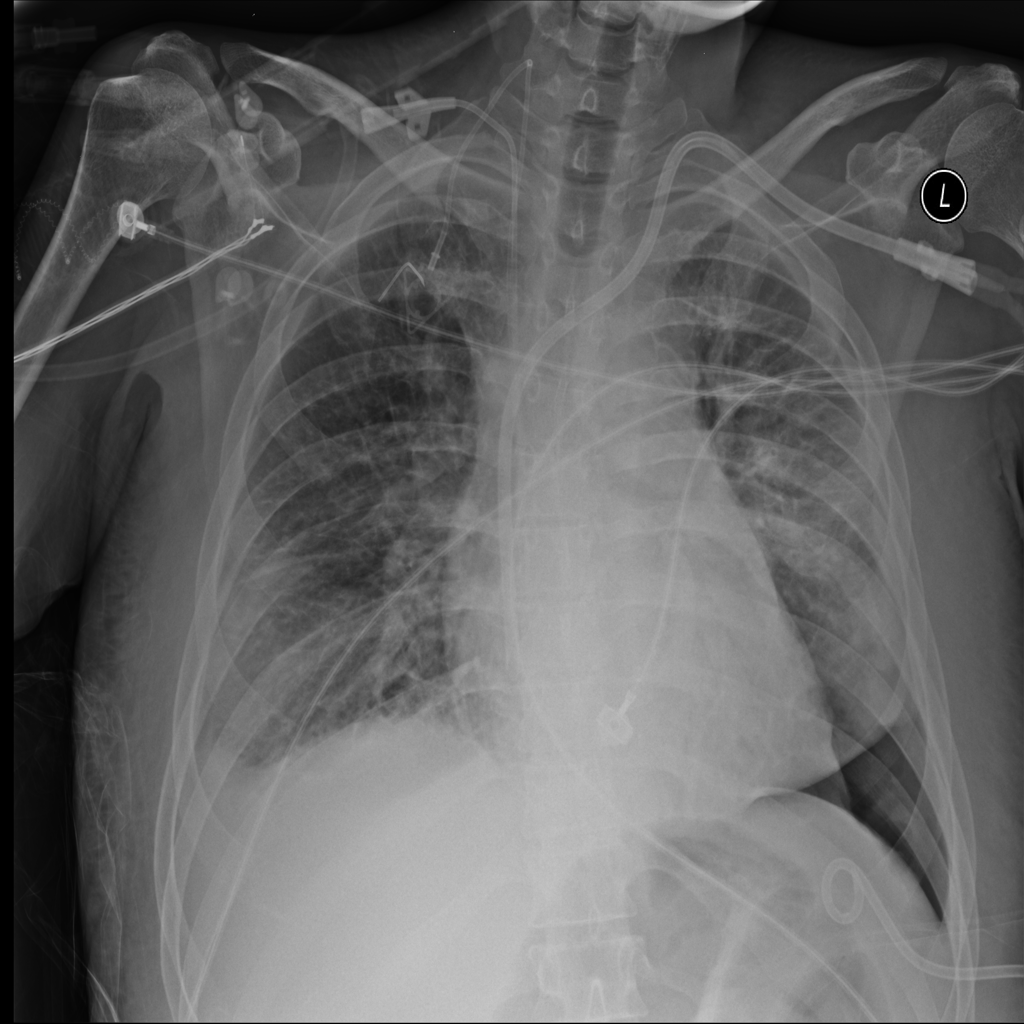}}   &  
              \frame{\includegraphics[width=0.155\textwidth,height=0.155\textwidth]{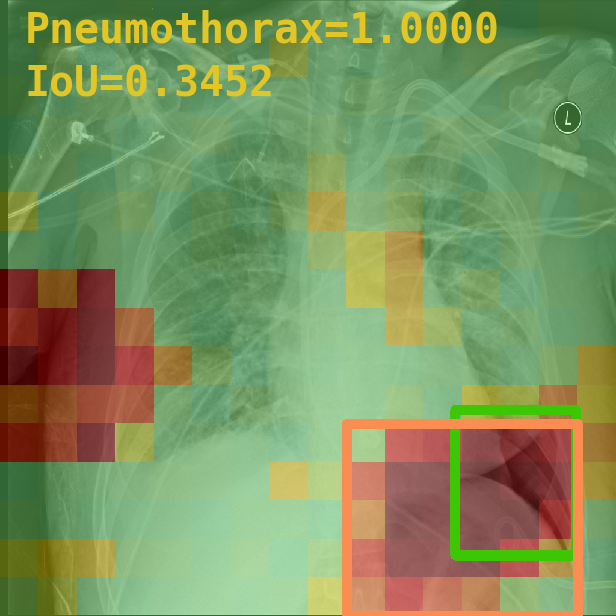}}   & 
              \frame{\includegraphics[width=0.155\textwidth,height=0.155\textwidth]{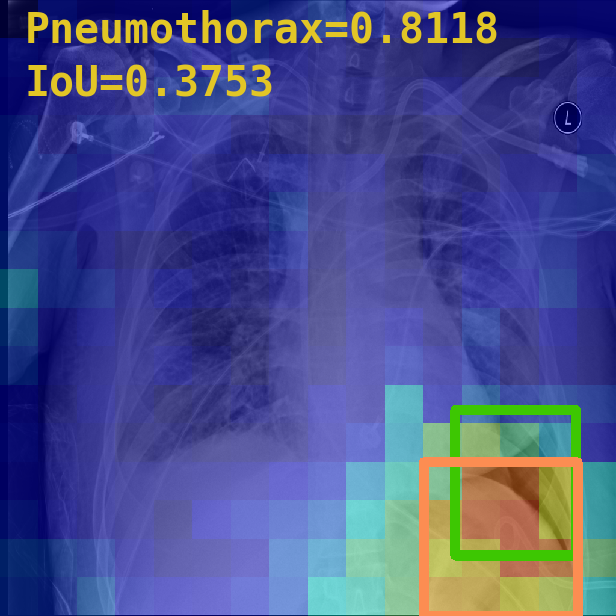}}  
               
        \end{tabular} 
        &   
        \begin{tabular}{ccc}
        \multicolumn{3}{c}{\fontfamily{lmtt}\selectfont\scriptsize Pneumonia}\\
              \frame{\includegraphics[width=0.155\textwidth,height=0.155\textwidth]{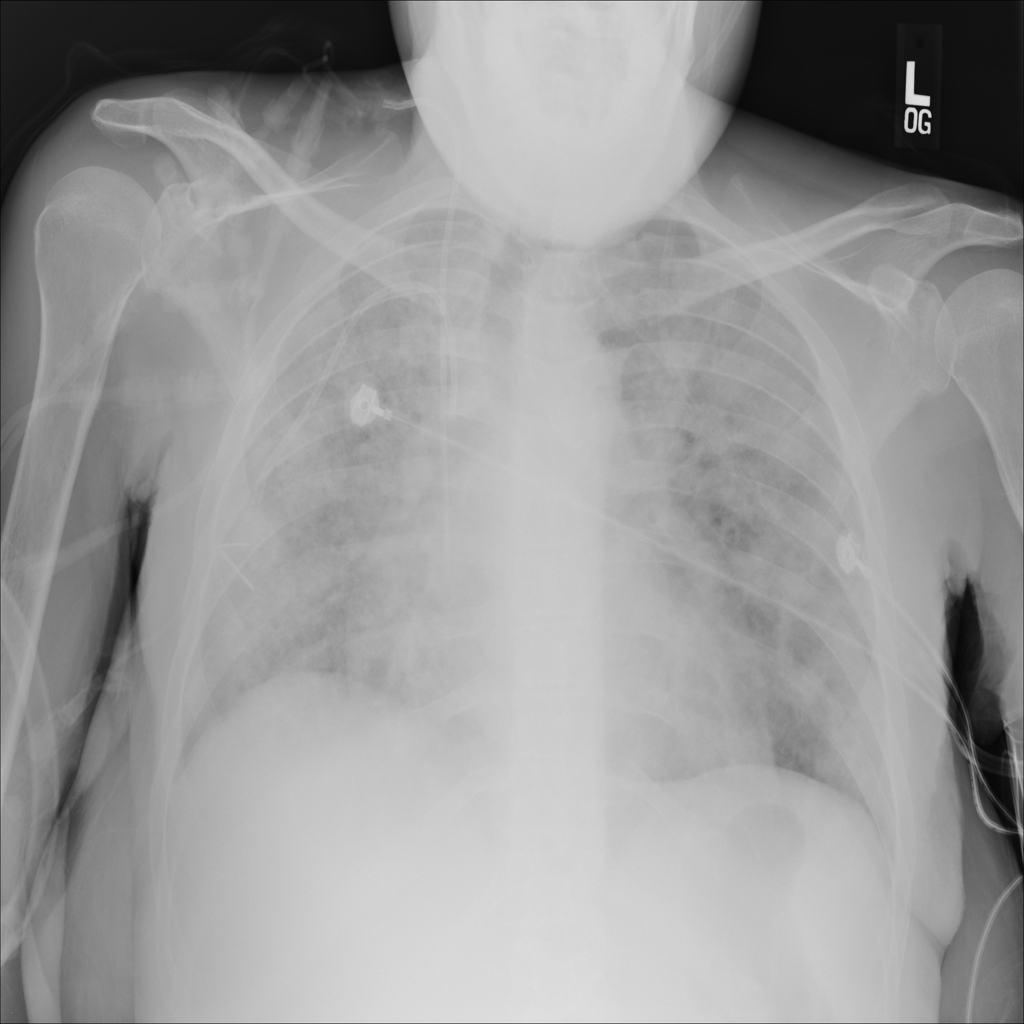}}   &  
              \frame{\includegraphics[width=0.155\textwidth,height=0.155\textwidth]{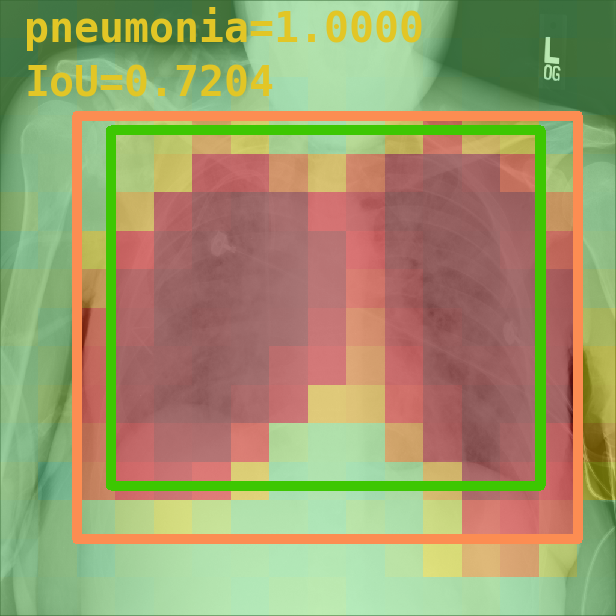}}   & 
              \frame{\includegraphics[width=0.155\textwidth,height=0.155\textwidth]{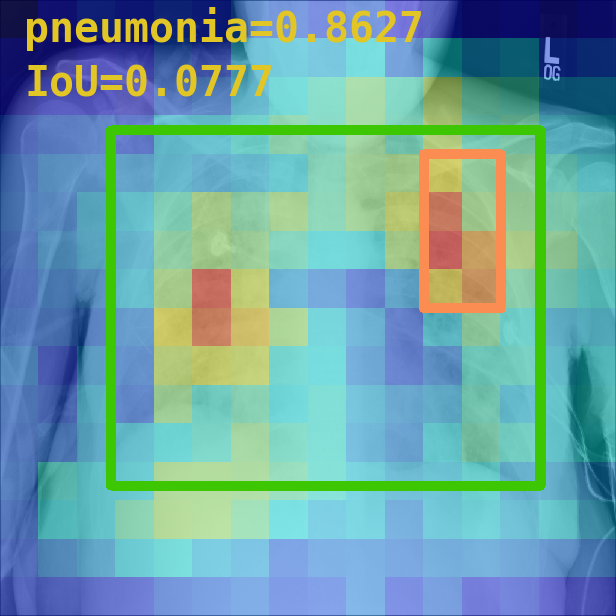}}   
             
        \end{tabular}\\ 
          \begin{tabular}{ccc}
          \multicolumn{3}{c}{\fontfamily{lmtt}\selectfont\scriptsize Mass}\\
              \frame{\includegraphics[width=0.155\textwidth,height=0.155\textwidth]{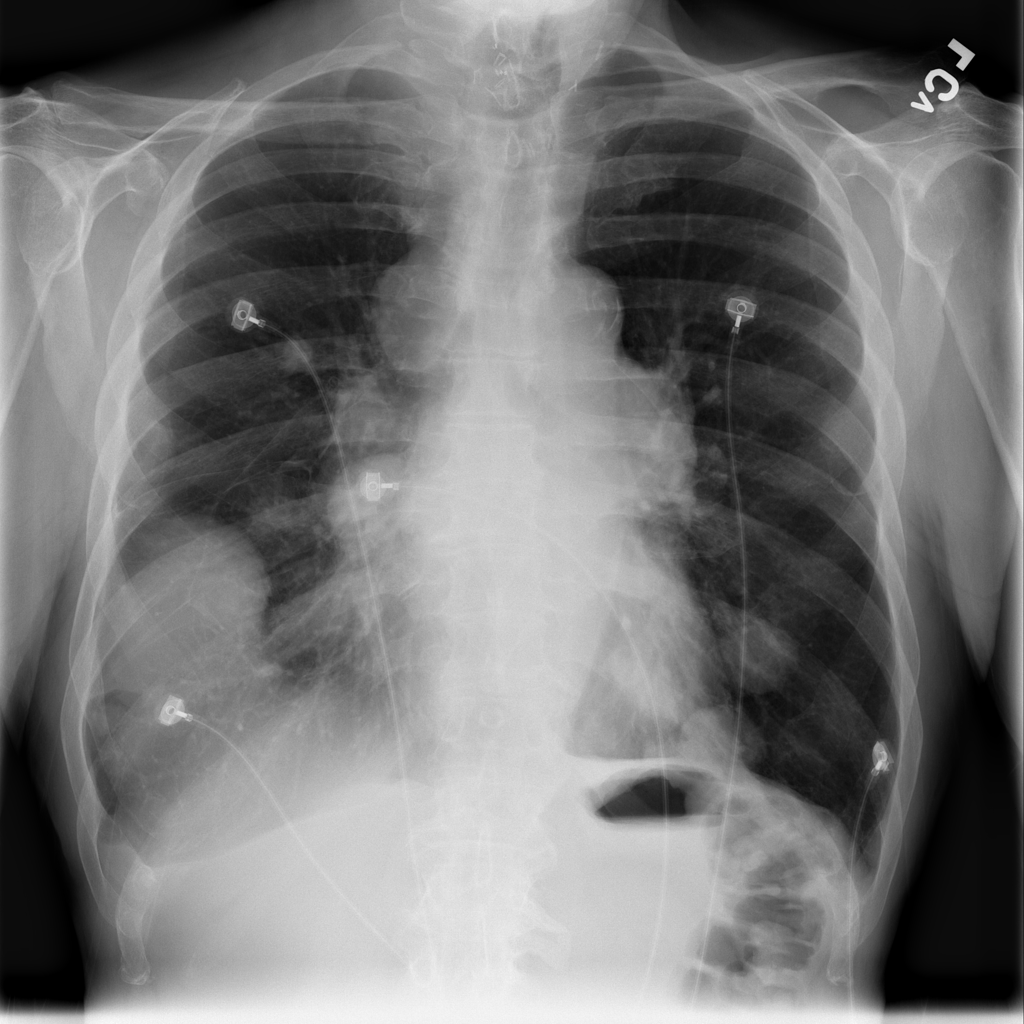}}   &  
              \frame{\includegraphics[width=0.155\textwidth,height=0.155\textwidth]{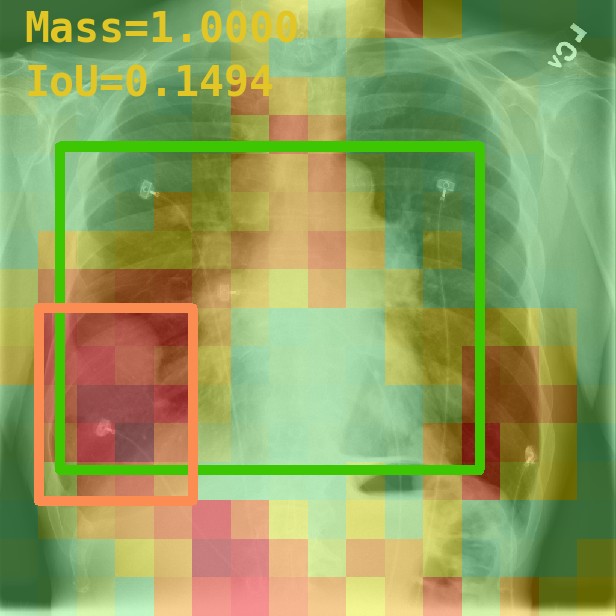}}   & 
              \frame{\includegraphics[width=0.155\textwidth,height=0.155\textwidth]{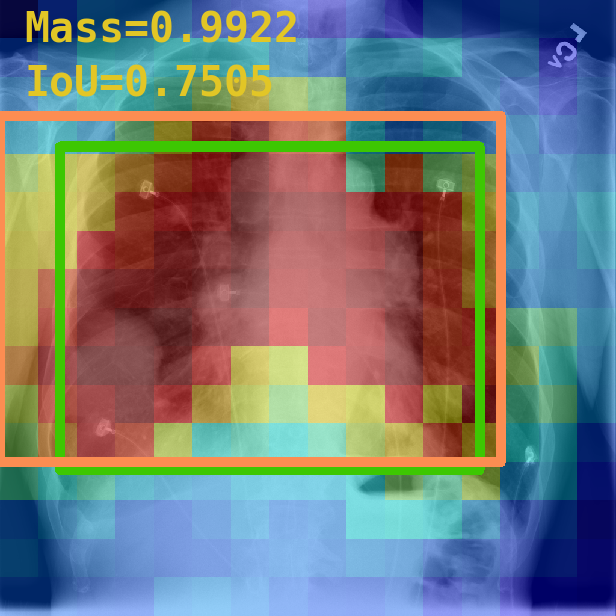}}

        \end{tabular}
        & 
         \begin{tabular}{ccc}
         \multicolumn{3}{c}{\fontfamily{lmtt}\selectfont\scriptsize Atelectasis}\\
              \frame{\includegraphics[width=0.155\textwidth,height=0.155\textwidth]{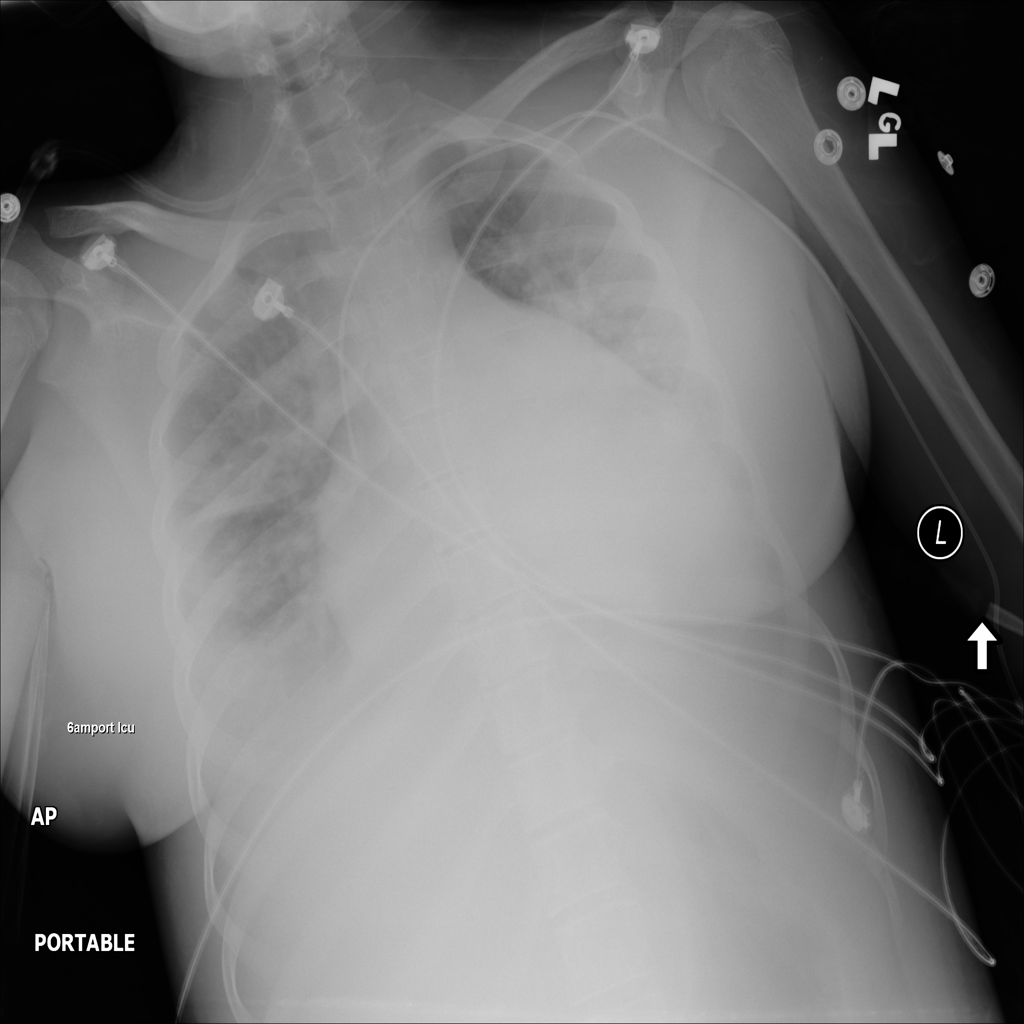}}   &  
              \frame{\includegraphics[width=0.155\textwidth,height=0.155\textwidth]{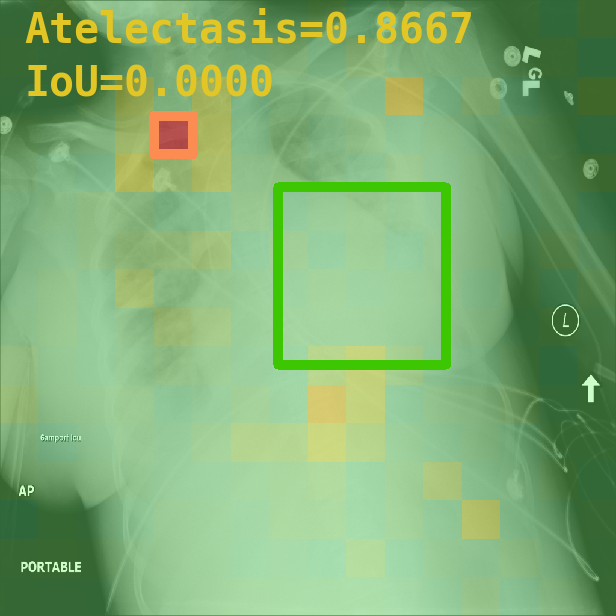}}   & 
              \frame{\includegraphics[width=0.155\textwidth,height=0.155\textwidth]{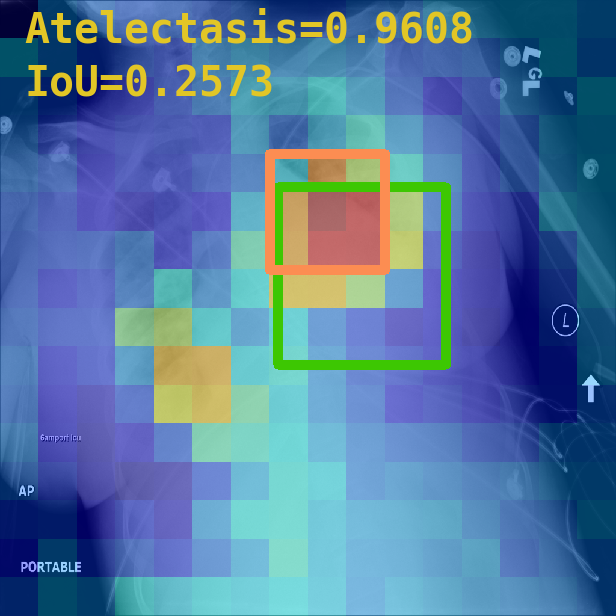}}   
             
        \end{tabular}\\ 
         \begin{tabular}{ccc}
         \multicolumn{3}{c}{\fontfamily{lmtt}\selectfont\scriptsize Nodule}\\
\includegraphics[width=0.155\textwidth,height=0.155\textwidth]{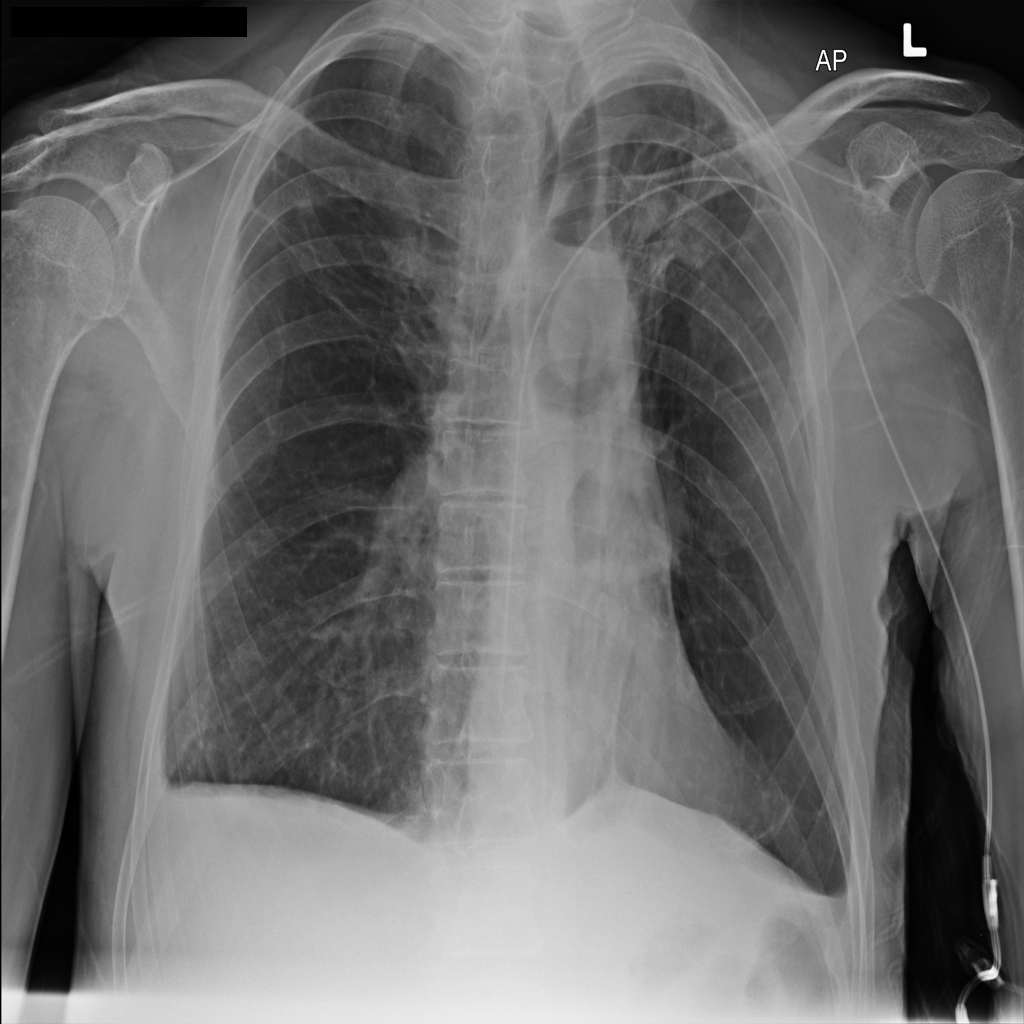}   &  
              \frame{\includegraphics[width=0.155\textwidth,height=0.155\textwidth]{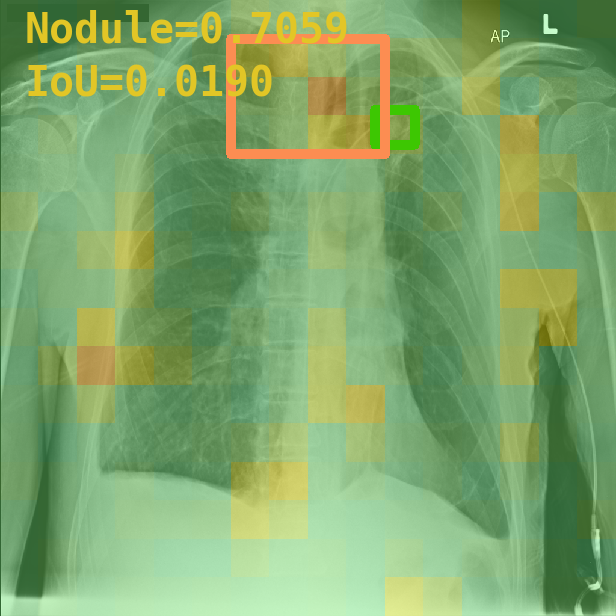}}   & 
             
             \frame{\includegraphics[width=0.155\textwidth,height=0.155\textwidth]{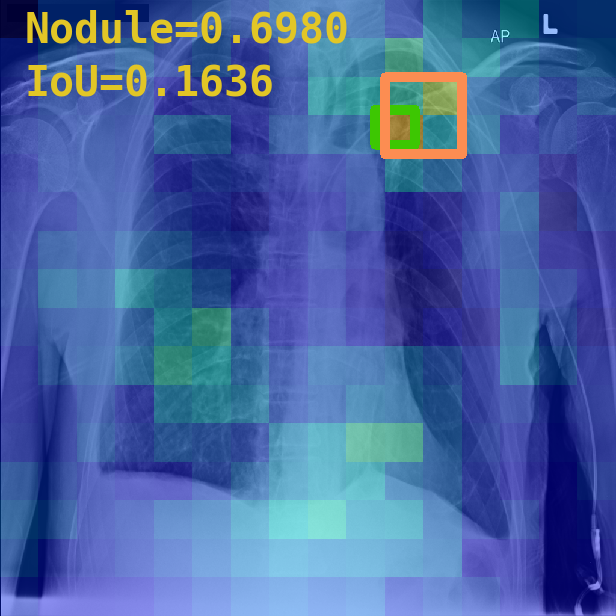}}   
        \end{tabular} 
        & 
         \begin{tabular}{ccc}
         \multicolumn{3}{c}{\fontfamily{lmtt}\selectfont\scriptsize Infiltrate}\\
              \frame{\includegraphics[width=0.155\textwidth,height=0.155\textwidth]{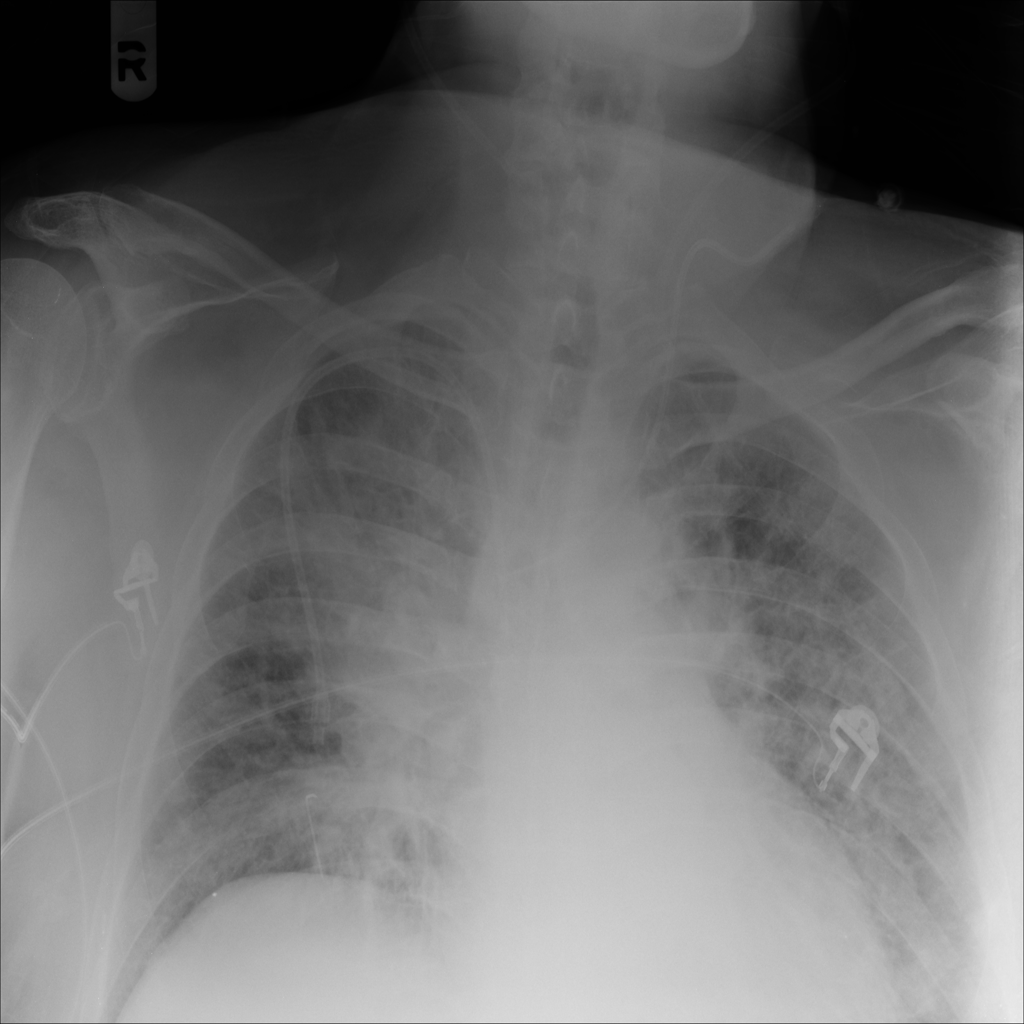}}   &  
              \frame{\includegraphics[width=0.155\textwidth,height=0.155\textwidth]{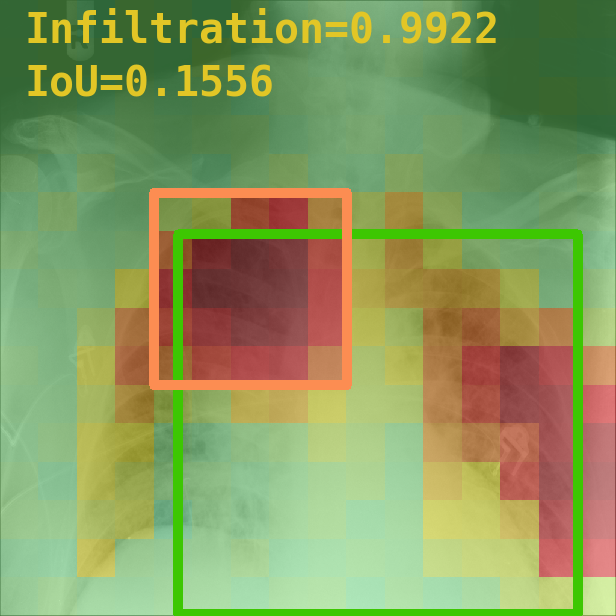}}   & 
             \frame{\includegraphics[width=0.155\textwidth,height=0.155\textwidth]{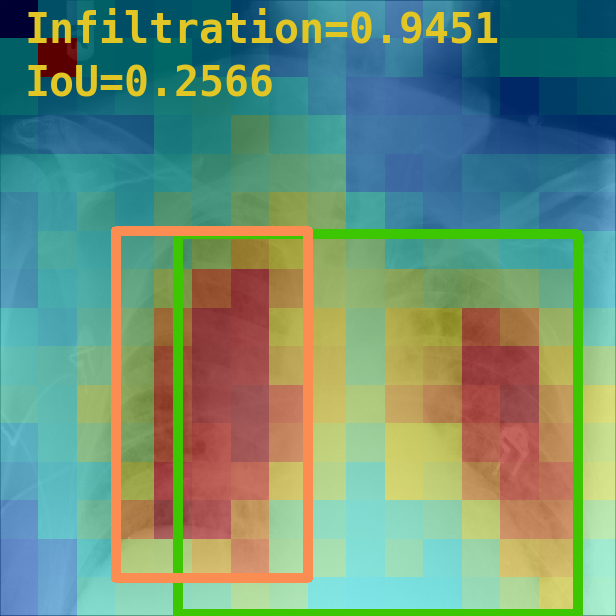}}      
              
        \end{tabular}\\ 
         \begin{tabular}{ccc}
         \multicolumn{3}{c}{\fontfamily{lmtt}\selectfont\scriptsize Effusion}\\
              \frame{\includegraphics[width=0.155\textwidth,height=0.155\textwidth]{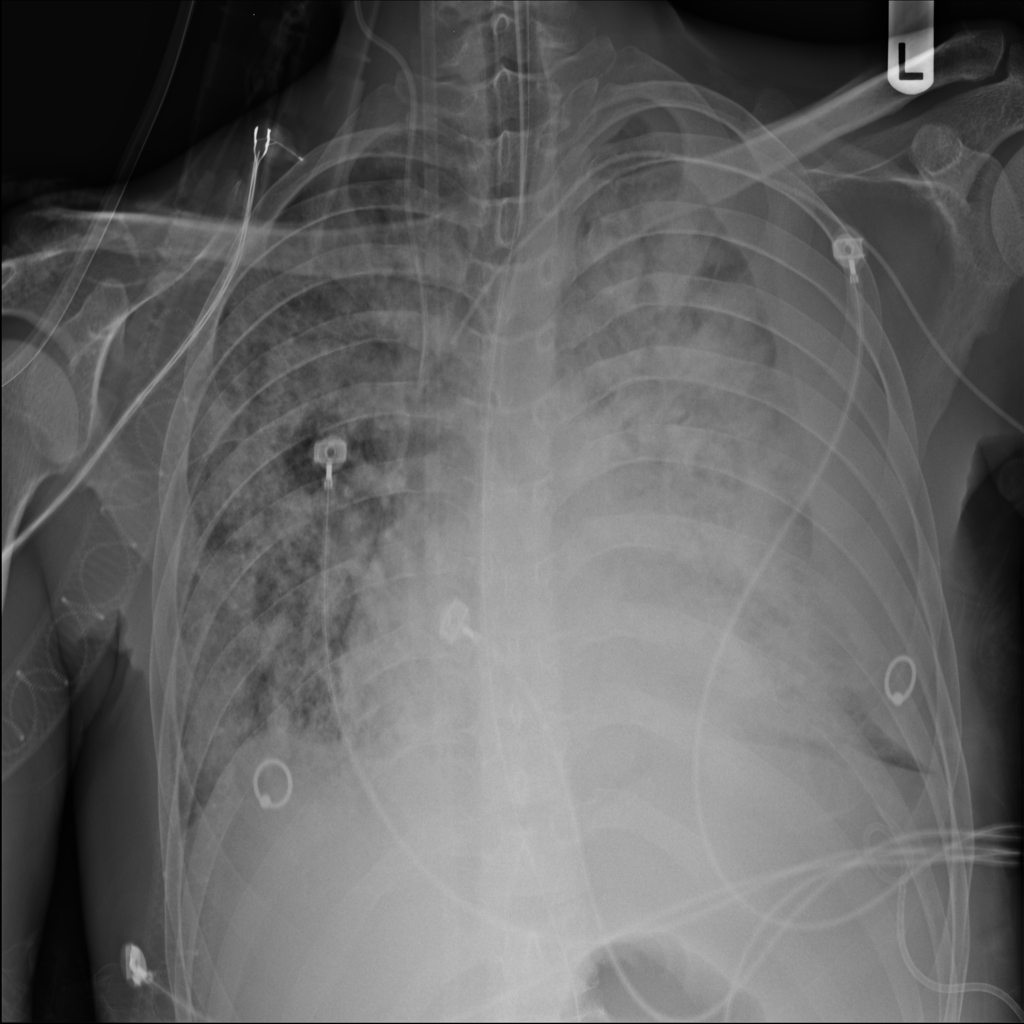}}   &  
              \frame{\includegraphics[width=0.155\textwidth,height=0.155\textwidth]{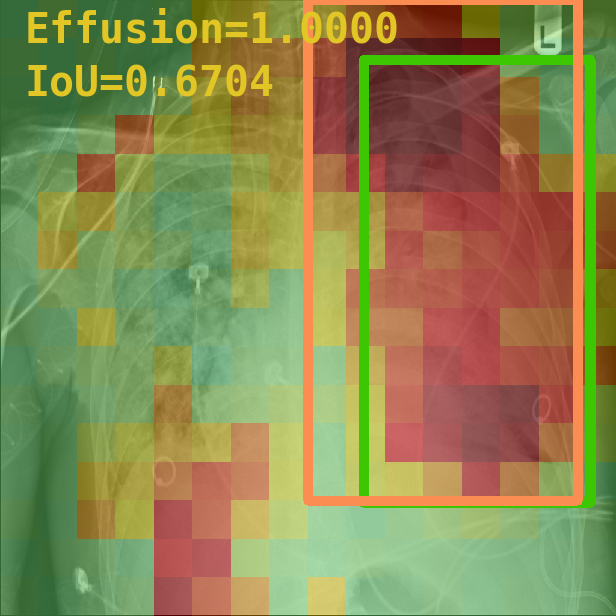}}   & 
              \frame{\includegraphics[width=0.155\textwidth,height=0.155\textwidth]{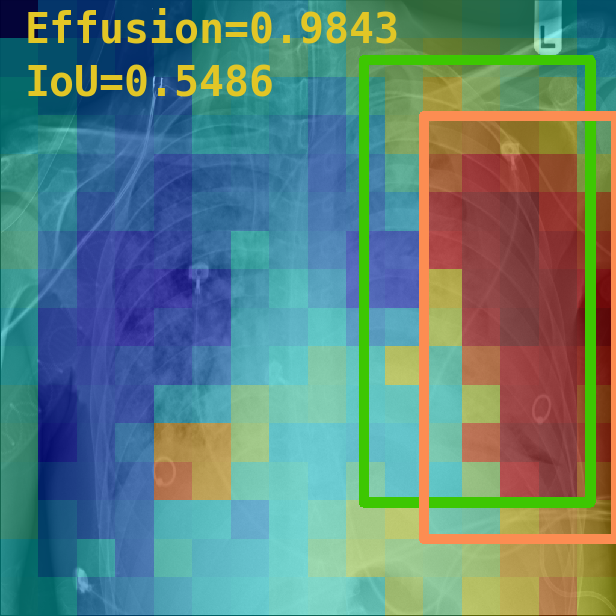}}   
              \\\fontfamily{lmtt}\selectfont\scriptsize  Input &\fontfamily{lmtt}\selectfont\scriptsize  Baseline & \fontfamily{lmtt}\selectfont\scriptsize SGL 
        \end{tabular}
        &
         \begin{tabular}{ccc}
         \multicolumn{3}{c}{\fontfamily{lmtt}\selectfont\scriptsize Cardiomegaly}\\
              \frame{\includegraphics[width=0.155\textwidth,height=0.155\textwidth]{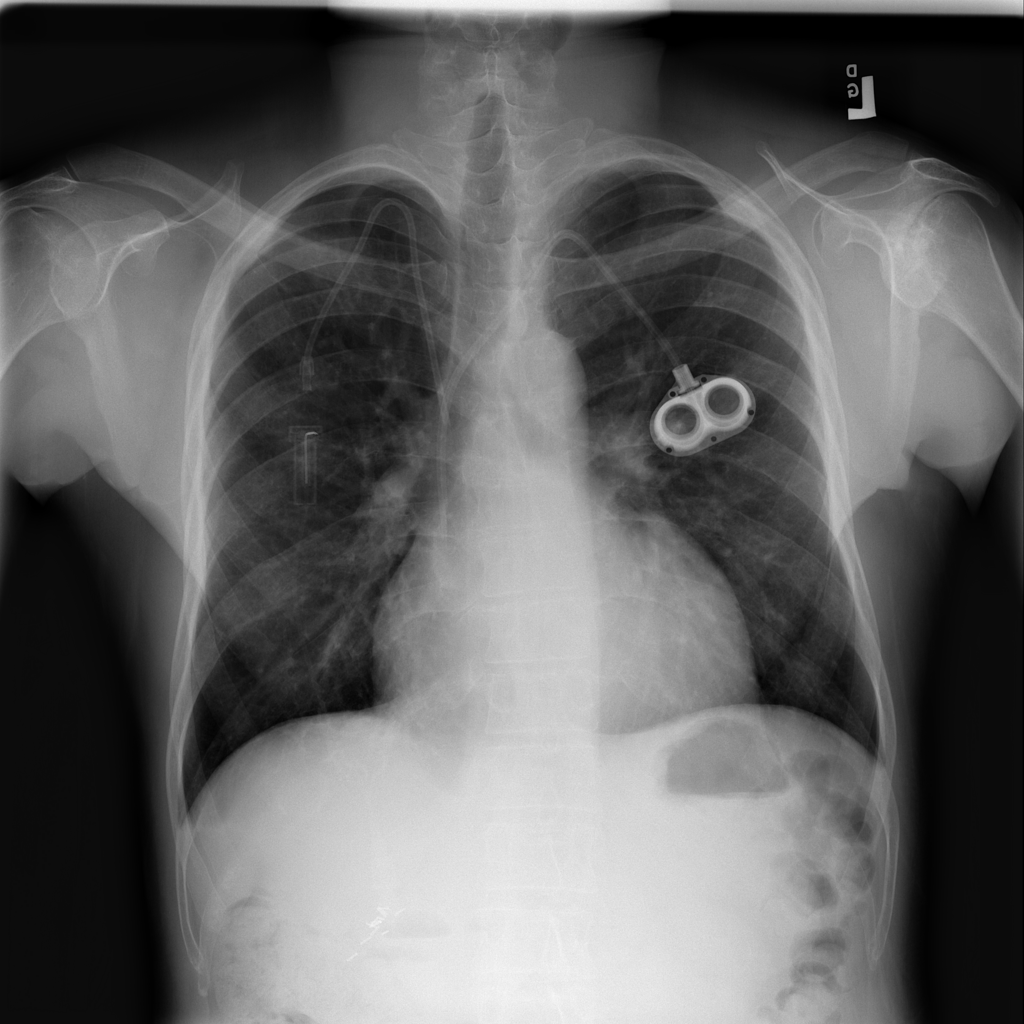}}   &  
              \frame{\includegraphics[width=0.155\textwidth,height=0.155\textwidth]{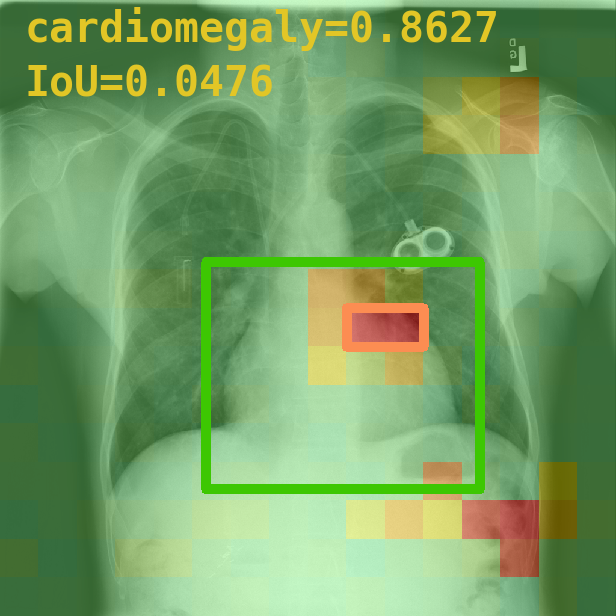}}   & 
              \frame{\includegraphics[width=0.155\textwidth,height=0.155\textwidth]{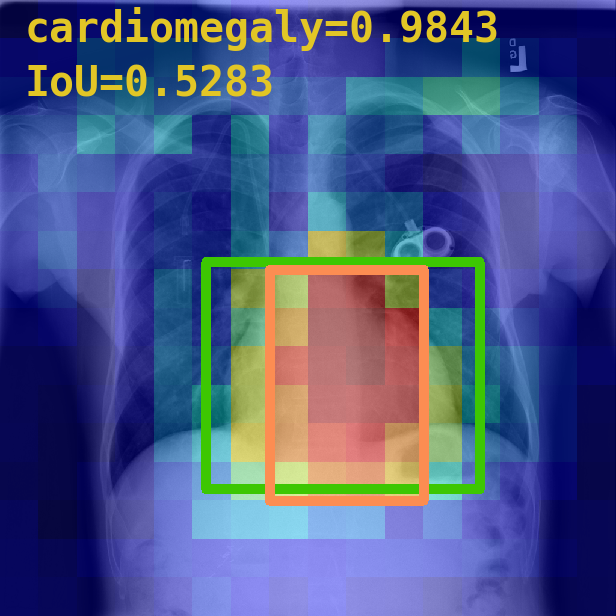}}   
              \\ \fontfamily{lmtt}\selectfont\scriptsize Input &\fontfamily{lmtt}\selectfont\scriptsize Baseline &\fontfamily{lmtt}\selectfont\scriptsize SGL 
        \end{tabular}
        
    \end{tabular}
    \caption{ We compare the patch-wise predictions between a mean-pooling trained baseline to our proposed method for different diseases. The value ranges from 0 (\textcolor{blue}{blue}) to 1 (\textcolor{red}{red}). We show prediction boxes (\textcolor{orange}{orange}) around the connected component of the maximum prediction and Ground-Truth bounding boxes (\textcolor{green}{green}).  }
    \label{fig:pathologies}
\end{figure}

\section{Conclusion}
In this paper, we propose a novel loss formulation in which one gathers auxiliary supervision from the network's predictions to provide instance-level supervision. 
In comparison to existing MIL-based loss functions, we do not rely on initialization and still provide pixel-wise supervision driving the network. Due to the design of this loss, it can support any MIL-setting such as patch-based pathology diagnosis. 
We demonstrate our method on two MIL-based datasets as well as the challenging NIH ChestX-Ray14 dataset. We display promising classification and localization performance qualitatively and quantitatively. 

\\\\
\noindent\textbf{Acknowledgements.}
The present contribution is supported by the Helmholtz Association under the joint research school ``HIDSS4Health – Helmholtz Information and Data Science School for Health''.

\clearpage
\bibliographystyle{splncs}
\bibliography{egbib}

\begin{thebibliography}{10}

\bibitem{nhs}
:
\newblock Nhs england: Diagnostic imaging dataset statistical release.
\newblock \url{https://www.england.nhs.uk/} (2020)

\bibitem{openi}
Kohli~MD, R.M.:
\newblock Open-i: Indiana university chest x-ray collection.
\newblock \url{https://openi.nlm.nih.gov} (2013)

\bibitem{bustos2019padchest}
Bustos, A., Pertusa, A., Salinas, J.M., de~la Iglesia-Vay{\'a}, M.:
\newblock Padchest: A large chest x-ray image dataset with multi-label
  annotated reports.
\newblock arXiv preprint arXiv:1901.07441 (2019)

\bibitem{johnson2019mimic}
Johnson, A.E., Pollard, T.J., Berkowitz, S., Greenbaum, N.R., Lungren, M.P.,
  Deng, C.y., Mark, R.G., Horng, S.:
\newblock Mimic-cxr: a large publicly available database of labeled chest
  radiographs.
\newblock arXiv preprint arXiv:1901.07042 \textbf{1} (2019)

\bibitem{irvin2019chexpert}
Irvin, J., Rajpurkar, P., Ko, M., Yu, Y., Ciurea-Ilcus, S., Chute, C.,
  Marklund, H., Haghgoo, B., Ball, R., Shpanskaya, K.,  et~al.:
\newblock Chexpert: A large chest radiograph dataset with uncertainty labels
  and expert comparison.
\newblock In: Proceedings of the AAAI Conference on Artificial Intelligence.
  Volume~33. (2019)  590--597

\bibitem{wang2017chestx}
Wang, X., Peng, Y., Lu, L., Lu, Z., Bagheri, M., Summers, R.M.:
\newblock Chestx-ray8: Hospital-scale chest x-ray database and benchmarks on
  weakly-supervised classification and localization of common thorax diseases.
\newblock In: Proceedings of the IEEE conference on computer vision and pattern
  recognition. (2017)  2097--2106

\bibitem{cai2018iterative}
Cai, J., Lu, L., Harrison, A.P., Shi, X., Chen, P., Yang, L.:
\newblock Iterative attention mining for weakly supervised thoracic disease
  pattern localization in chest x-rays.
\newblock In: International Conference on Medical Image Computing and
  Computer-Assisted Intervention, Springer (2018)  589--598

\bibitem{shen2018dynamic}
Shen, Y., Gao, M.:
\newblock Dynamic routing on deep neural network for thoracic disease
  classification and sensitive area localization.
\newblock In: International Workshop on Machine Learning in Medical Imaging,
  Springer (2018)  389--397

\bibitem{tang2018attention}
Tang, Y., Wang, X., Harrison, A.P., Lu, L., Xiao, J., Summers, R.M.:
\newblock Attention-guided curriculum learning for weakly supervised
  classification and localization of thoracic diseases on chest radiographs.
\newblock In: International Workshop on Machine Learning in Medical Imaging,
  Springer (2018)  249--258

\bibitem{baltruschat2019comparison}
Baltruschat, I.M., Nickisch, H., Grass, M., Knopp, T., Saalbach, A.:
\newblock Comparison of deep learning approaches for multi-label chest x-ray
  classification.
\newblock Scientific reports \textbf{9} (2019)  1--10

\bibitem{rajpurkar2017chexnet}
Rajpurkar, P., Irvin, J., Zhu, K., Yang, B., Mehta, H., Duan, T., Ding, D.,
  Bagul, A., Langlotz, C., Shpanskaya, K.,  et~al.:
\newblock Chexnet: Radiologist-level pneumonia detection on chest x-rays with
  deep learning.
\newblock arXiv preprint arXiv:1711.05225 (2017)

\bibitem{wang2018chestnet}
Wang, H., Xia, Y.:
\newblock Chestnet: A deep neural network for classification of thoracic
  diseases on chest radiography.
\newblock arXiv preprint arXiv:1807.03058 (2018)

\bibitem{park2019curriculum}
Park, B., Cho, Y., Lee, G., Lee, S.M., Cho, Y.H., Lee, E.S., Lee, K.H., Seo,
  J.B., Kim, N.:
\newblock A curriculum learning strategy to enhance the accuracy of
  classification of various lesions in chest-pa x-ray screening for pulmonary
  abnormalities.
\newblock Scientific reports \textbf{9} (2019)  1--9

\bibitem{rajpurkar2018deep}
Rajpurkar, P., Irvin, J., Ball, R.L., Zhu, K., Yang, B., Mehta, H., Duan, T.,
  Ding, D., Bagul, A., Langlotz, C.P.,  et~al.:
\newblock Deep learning for chest radiograph diagnosis: A retrospective
  comparison of the chexnext algorithm to practicing radiologists.
\newblock PLoS medicine \textbf{15} (2018)  e1002686

\bibitem{wang2018low}
Wang, Q., Cheng, J.Z., Zhou, Y., Zhuang, H., Li, C., Chen, B., Liu, Z., Huang,
  J., Wang, C., Zhou, X.:
\newblock Low-shot multi-label incremental learning for thoracic diseases
  diagnosis.
\newblock In: International Conference on Neural Information Processing,
  Springer (2018)  420--432

\bibitem{li2019knowledge}
Li, C.Y., Liang, X., Hu, Z., Xing, E.P.:
\newblock Knowledge-driven encode, retrieve, paraphrase for medical image
  report generation.
\newblock In: Proceedings of the AAAI Conference on Artificial Intelligence.
  Volume~33. (2019)  6666--6673

\bibitem{li2019vispi}
Li, X., Cao, R., Zhu, D.:
\newblock Vispi: Automatic visual perception and interpretation of chest
  x-rays.
\newblock arXiv preprint arXiv:1906.05190 (2019)

\bibitem{li2020netnet}
Li, Y., Pang, Y., Shen, J., Cao, J., Shao, L.:
\newblock Netnet: Neighbor erasing and transferring network for better single
  shot object detection.
\newblock arXiv preprint arXiv:2001.06690 (2020)

\bibitem{wang2018tienet}
Wang, X., Peng, Y., Lu, L., Lu, Z., Summers, R.M.:
\newblock Tienet: Text-image embedding network for common thorax disease
  classification and reporting in chest x-rays.
\newblock In: Proceedings of the IEEE conference on computer vision and pattern
  recognition. (2018)  9049--9058

\bibitem{yan2018weakly}
Yan, C., Yao, J., Li, R., Xu, Z., Huang, J.:
\newblock Weakly supervised deep learning for thoracic disease classification
  and localization on chest x-rays.
\newblock In: Proceedings of the 2018 ACM International Conference on
  Bioinformatics, Computational Biology, and Health Informatics. (2018)
  103--110

\bibitem{li2018thoracic}
Li, Z., Wang, C., Han, M., Xue, Y., Wei, W., Li, L.J., Fei-Fei, L.:
\newblock Thoracic disease identification and localization with limited
  supervision.
\newblock In: Proceedings of the IEEE Conference on Computer Vision and Pattern
  Recognition. (2018)  8290--8299

\bibitem{liu2019align}
Liu, J., Zhao, G., Fei, Y., Zhang, M., Wang, Y., Yu, Y.:
\newblock Align, attend and locate: Chest x-ray diagnosis via contrast induced
  attention network with limited supervision.
\newblock In: Proceedings of the IEEE International Conference on Computer
  Vision. (2019)  10632--10641

\bibitem{rozenberglocalization}
Rozenberg, E., Freedman, D., Bronstein, A.:
\newblock Localization with limited annotation for chest x-rays.
\newblock (2019)

\bibitem{guan2018multi}
Guan, Q., Huang, Y.:
\newblock Multi-label chest x-ray image classification via category-wise
  residual attention learning.
\newblock Pattern Recognition Letters (2018)

\bibitem{ren2015faster}
Ren, S., He, K., Girshick, R., Sun, J.:
\newblock Faster r-cnn: Towards real-time object detection with region proposal
  networks.
\newblock In: Advances in neural information processing systems. (2015)  91--99

\bibitem{liu2016ssd}
Liu, W., Anguelov, D., Erhan, D., Szegedy, C., Reed, S., Fu, C.Y., Berg, A.C.:
\newblock Ssd: Single shot multibox detector.
\newblock In: European conference on computer vision, Springer (2016)  21--37

\bibitem{redmon2018yolov3}
Redmon, J., Farhadi, A.:
\newblock Yolov3: An incremental improvement.
\newblock arXiv preprint arXiv:1804.02767 (2018)

\bibitem{zhou2016learning}
Zhou, B., Khosla, A., Lapedriza, A., Oliva, A., Torralba, A.:
\newblock Learning deep features for discriminative localization.
\newblock In: Proceedings of the IEEE conference on computer vision and pattern
  recognition. (2016)  2921--2929

\bibitem{selvaraju2017grad}
Selvaraju, R.R., Cogswell, M., Das, A., Vedantam, R., Parikh, D., Batra, D.:
\newblock Grad-cam: Visual explanations from deep networks via gradient-based
  localization.
\newblock In: Proceedings of the IEEE international conference on computer
  vision. (2017)  618--626

\bibitem{zhang2018top}
Zhang, J., Bargal, S.A., Lin, Z., Brandt, J., Shen, X., Sclaroff, S.:
\newblock Top-down neural attention by excitation backprop.
\newblock International Journal of Computer Vision \textbf{126} (2018)
  1084--1102

\bibitem{yao2017learning}
Yao, L., Poblenz, E., Dagunts, D., Covington, B., Bernard, D., Lyman, K.:
\newblock Learning to diagnose from scratch by exploiting dependencies among
  labels.
\newblock arXiv preprint arXiv:1710.10501 (2017)

\bibitem{ilse2018attention}
Ilse, M., Tomczak, J.M., Welling, M.:
\newblock Attention-based deep multiple instance learning.
\newblock arXiv preprint arXiv:1802.04712 (2018)

\bibitem{zhou2017adaptive}
Zhou, Y., Sun, X., Liu, D., Zha, Z., Zeng, W.:
\newblock Adaptive pooling in multi-instance learning for web video annotation.
\newblock In: Proceedings of the IEEE International Conference on Computer
  Vision Workshops. (2017)  318--327

\bibitem{mcfee2018adaptive}
McFee, B., Salamon, J., Bello, J.P.:
\newblock Adaptive pooling operators for weakly labeled sound event detection.
\newblock IEEE/ACM Transactions on Audio, Speech, and Language Processing
  \textbf{26} (2018)  2180--2193

\bibitem{wang2019comparison}
Wang, Y., Li, J., Metze, F.:
\newblock A comparison of five multiple instance learning pooling functions for
  sound event detection with weak labeling.
\newblock In: ICASSP 2019-2019 IEEE International Conference on Acoustics,
  Speech and Signal Processing (ICASSP), IEEE (2019)  31--35

\bibitem{liao2019evaluate}
Liao, F., Liang, M., Li, Z., Hu, X., Song, S.:
\newblock Evaluate the malignancy of pulmonary nodules using the 3-d deep leaky
  noisy-or network.
\newblock IEEE transactions on neural networks and learning systems \textbf{30}
  (2019)  3484--3495

\bibitem{yan2018deep}
Yan, Y., Wang, X., Guo, X., Fang, J., Liu, W., Huang, J.:
\newblock Deep multi-instance learning with dynamic pooling.
\newblock In: Asian Conference on Machine Learning. (2018)  662--677

\bibitem{chen2019deep}
Chen, H., Miao, S., Xu, D., Hager, G.D., Harrison, A.P.:
\newblock Deep hierarchical multi-label classification of chest x-ray images.
\newblock Proceedings of Machine Learning Research \textbf{1} (2019) ~13

\bibitem{yao2018weakly}
Yao, L., Prosky, J., Poblenz, E., Covington, B., Lyman, K.:
\newblock Weakly supervised medical diagnosis and localization from multiple
  resolutions.
\newblock arXiv preprint arXiv:1803.07703 (2018)

\bibitem{guendel2019multi}
Guendel, S., Ghesu, F.C., Grbic, S., Gibson, E., Georgescu, B., Maier, A.,
  Comaniciu, D.:
\newblock Multi-task learning for chest x-ray abnormality classification on
  noisy labels.
\newblock arXiv preprint arXiv:1905.06362 (2019)

\bibitem{liu2019clinically}
Liu, G., Hsu, T.M.H., McDermott, M., Boag, W., Weng, W.H., Szolovits, P.,
  Ghassemi, M.:
\newblock Clinically accurate chest x-ray report generation.
\newblock arXiv preprint arXiv:1904.02633 (2019)

\bibitem{zhang2018self}
Zhang, X., Wei, Y., Kang, G., Yang, Y., Huang, T.:
\newblock Self-produced guidance for weakly-supervised object localization.
\newblock In: Proceedings of the European Conference on Computer Vision (ECCV).
  (2018)  597--613

\bibitem{zhang2018adversarial}
Zhang, X., Wei, Y., Feng, J., Yang, Y., Huang, T.S.:
\newblock Adversarial complementary learning for weakly supervised object
  localization.
\newblock In: Proceedings of the IEEE Conference on Computer Vision and Pattern
  Recognition. (2018)  1325--1334

\bibitem{wei2017object}
Wei, Y., Feng, J., Liang, X., Cheng, M.M., Zhao, Y., Yan, S.:
\newblock Object region mining with adversarial erasing: A simple
  classification to semantic segmentation approach.
\newblock In: Proceedings of the IEEE conference on computer vision and pattern
  recognition. (2017)  1568--1576

\bibitem{dietterich1997solving}
Dietterich, T.G., Lathrop, R.H., Lozano-P{\'e}rez, T.:
\newblock Solving the multiple instance problem with axis-parallel rectangles.
\newblock Artificial intelligence \textbf{89} (1997)  31--71

\bibitem{kong2019sound}
Kong, Q., Xu, Y., Sobieraj, I., Wang, W., Plumbley, M.D.:
\newblock Sound event detection and time--frequency segmentation from weakly
  labelled data.
\newblock IEEE/ACM Transactions on Audio, Speech, and Language Processing
  \textbf{27} (2019)  777--787

\bibitem{tang2017multiple}
Tang, P., Wang, X., Bai, X., Liu, W.:
\newblock Multiple instance detection network with online instance classifier
  refinement.
\newblock In: Proceedings of the IEEE Conference on Computer Vision and Pattern
  Recognition. (2017)  2843--2851

\bibitem{cinbis2016weakly}
Cinbis, R.G., Verbeek, J., Schmid, C.:
\newblock Weakly supervised object localization with multi-fold multiple
  instance learning.
\newblock IEEE transactions on pattern analysis and machine intelligence
  \textbf{39} (2016)  189--203

\bibitem{felipe2020distilling}
Felipe~Zeni, L., Jung, C.R.:
\newblock Distilling knowledge from refinement in multiple instance detection
  networks.
\newblock In: Proceedings of the IEEE/CVF Conference on Computer Vision and
  Pattern Recognition Workshops. (2020)  768--769

\bibitem{morfi2018data}
Morfi, V., Stowell, D.:
\newblock Data-efficient weakly supervised learning for low-resource audio
  event detection using deep learning.
\newblock arXiv preprint arXiv:1807.06972 (2018)

\bibitem{wang2015relaxed}
Wang, X., Zhu, Z., Yao, C., Bai, X.:
\newblock Relaxed multiple-instance svm with application to object discovery.
\newblock In: Proceedings of the IEEE International Conference on Computer
  Vision. (2015)  1224--1232

\bibitem{shamsolmoali2020amil}
Shamsolmoali, P., Zareapoor, M., Zhou, H., Yang, J.:
\newblock Amil: Adversarial multi-instance learning for human pose estimation.
\newblock ACM Transactions on Multimedia Computing, Communications, and
  Applications (TOMM) \textbf{16} (2020)  1--23

\bibitem{Hou2018Self}
Hou, Q., Jiang, P., Wei, Y., Cheng, M.M.:
\newblock Self-erasing network for integral object attention.
\newblock In: Advances in Neural Information Processing Systems. (2018)
  549--559

\bibitem{szegedy2016rethinking}
Szegedy, C., Vanhoucke, V., Ioffe, S., Shlens, J., Wojna, Z.:
\newblock Rethinking the inception architecture for computer vision.
\newblock In: Proceedings of the IEEE conference on computer vision and pattern
  recognition. (2016)  2818--2826

\bibitem{lecun1998gradient}
LeCun, Y., Bottou, L., Bengio, Y., Haffner, P.:
\newblock Gradient-based learning applied to document recognition.
\newblock Proceedings of the IEEE \textbf{86} (1998)  2278--2324

\bibitem{krizhevsky2009learning}
Krizhevsky, A., Hinton, G.,  et~al.:
\newblock Learning multiple layers of features from tiny images.
\newblock (2009)

\bibitem{he2016deep}
He, K., Zhang, X., Ren, S., Sun, J.:
\newblock Deep residual learning for image recognition.
\newblock In: Proceedings of the IEEE conference on computer vision and pattern
  recognition. (2016)  770--778

\bibitem{russakovsky2015imagenet}
Russakovsky, O., Deng, J., Su, H., Krause, J., Satheesh, S., Ma, S., Huang, Z.,
  Karpathy, A., Khosla, A., Bernstein, M.,  et~al.:
\newblock Imagenet large scale visual recognition challenge.
\newblock International journal of computer vision \textbf{115} (2015)
  211--252

\bibitem{kingma2014adam}
Kingma, D.P., Ba, J.:
\newblock Adam: A method for stochastic optimization.
\newblock arXiv preprint arXiv:1412.6980 (2014)

\bibitem{paszke2017automatic}
Paszke, A., Gross, S., Chintala, S., Chanan, G., Yang, E., DeVito, Z., Lin, Z.,
  Desmaison, A., Antiga, L., Lerer, A.:
\newblock Automatic differentiation in pytorch.
\newblock In: NIPS-W. (2017)

\end{thebibliography}

\end{document}